\documentclass{CUP-JNL-DTM}

\usepackage{graphicx}
\usepackage{multicol,multirow}
\usepackage{amsmath,amssymb,amsfonts}
\usepackage{mathrsfs}
\usepackage{amsthm}
\usepackage{rotating}
\usepackage{appendix}
\usepackage[natbib,style=apa]{biblatex}
\usepackage{ifpdf}
\usepackage[T1]{fontenc}
\usepackage{newtxtext}
\usepackage{newtxmath}
\usepackage{textcomp}
\usepackage{xcolor}
\usepackage{lipsum}
\usepackage[]{hyperref}
\hypersetup{colorlinks,allcolors=blue}
\usepackage{caption}
\usepackage{subcaption}
\usepackage{multirow}
\usepackage{booktabs}
\usepackage{rotating,tabularx}
\usepackage{colortbl} % For adding colored cells to tables
\usepackage{makecell} % For adding linebreaks to cells
\usepackage{multirow}

\usepackage{booktabs}
\usepackage{rotating,tabularx}
\usepackage{colortbl} % For adding colored cells to tables
\usepackage{makecell} % For adding linebreaks to cells

\theoremstyle{definition}

\numberwithin{equation}{section}

\jname{Wearable Technologies}
\articletype{RESEARCH ARTICLE}
\jyear{YEAR}
\begin{document}

\begin{Frontmatter}

\title[Article Title]{Sensorless model-based tension control for a cable-driven exosuit}

% There is no need to include ORCID IDs in your .pdf; this information is captured by the submission portal when a manuscript is submitted. 
\author[1,2]{Elena Bardi\textsuperscript{$\dag$}}
\author[3]{Adrian Esser\textsuperscript{$\dag$}}
\author[3]{Peter Wolf}
\author[1,2]{Marta Gandolla}
\author[1,4]{Emilia Ambrosini}
\author[1,4]{Alessandra Pedrocchi}
\author[3]{Robert Riener}

\authormark{Bardi \textit{et al}.}

\address[]{\textsuperscript{$\dag$}These authors contributed equally to this work and share first authorship}

\address[1]{\orgdiv{WeCobot Lab, Polo territoriale di Lecco}, \orgname{Politecnico di Milano}, \orgaddress{\city{Milano}, \postcode{20133}, \country{Italy}}}

\address[2]{\orgdiv{Department of Mechanical Engineering}, \orgaddress{\city{Milano}, \postcode{20133}, \country{Italy}}}

\address[3]{\orgdiv{Sensory Motor Systems Lab, Department of Health Sciences and Technology}, \orgname{ETH Zürich}, \orgaddress{\city{Zürich}, \postcode{8092}, \country{Switzerland}}}

\address[4]{\orgdiv{NEARLab, Department of Electronics, Informatics and Bioengineering}, \orgname{Politecnico di Milano}, \orgaddress{\city{Milano}, \postcode{20133}, \country{Italy}}}

\address[]{\textsuperscript{*}Please address all correspondence to: \url{adrian.esser@hest.ethz.ch}}


%\keywords[MSC Codes]{\codes[Primary]{CODE1}; \codes[Secondary]{CODE2, CODE3}}

\abstract{  Cable-driven exosuits have the potential to support individuals with motor disabilities across the continuum of care. When supporting a limb with a cable, force sensors are often used to measure tension. However, force sensors add cost, complexity, and distal components. This paper presents a design and control approach to remove the force sensor from an upper limb cable-driven exosuit. A mechanical design for the exosuit was developed to maximize passive transparency. Then, a data-driven friction identification was conducted on a mannequin test bench to design a model-based tension controller. Seventeen healthy participants raised and lowered their right arms to evaluate tension tracking, movement quality, and muscular effort. Questionnaires on discomfort, physical exertion, and fatigue were collected. The proposed strategy allowed tracking the desired assistive torque with an RMSE of 0.71 Nm (18\%) at 50\% gravity support. During the raising phase, the EMG signals of the anterior deltoid, trapezius, and pectoralis major were reduced on average compared to the no-suit condition by 30\%, 38\%, and 38\%, respectively. The posterior deltoid activity was increased by 32\% during lowering. Position tracking was not significantly altered, whereas movement smoothness significantly decreased. This work demonstrates the feasibility and effectiveness of removing the force sensor from a cable-driven exosuit. A significant increase in discomfort in the lower neck and right shoulder indicated that the ergonomics of the suit could be improved. Overall this work paves the way towards simpler and more affordable exosuits.
}

\end{Frontmatter}

% Some math journals (FLO) require a table of contents. Comment out this line if no ToC is needed.
\localtableofcontents

\section{Introduction}
\subsection{Motivation of the study}
Impairments of the upper limb(s) may result from conditions such as stroke, spinal cord injury, muscular dystrophy, multiple sclerosis, amyotrophic lateral sclerosis, and ageing and might strongly affect the ability to conduct daily activities autonomously \citep{Natterlund2001, Bohannon2007, Wang2020}. Motor rehabilitation is fundamental in addressing the loss of any ability \citep{Faria-Fortini2011}. In cases where complete recovery is unattainable, individuals may require support with everyday activities through the use of assistive devices, which aim to restore independence to individuals with disabilities \citep{Longatelli2021, Readioff2022}.

In the field of assistive wearable robotics, there is a growing interest in soft wearable devices, referred to as exosuits \citep{Georgarakis2022, Noronha2022, Proietti2023}. Key features of exosuits include being lightweight, portable, comfortable, and less constraining for natural movements, making them suitable for daily use \citep{Xiloyannis2021, Bardi2022}. Among the possible exosuit types, such as pneumatic \citep{Das2020, Proietti2023}, DC motor-based cable-driven (or tendon-driven) exosuits are promising because of the maturity of the technology and the possibility of locating the control and actuation units proximally on the body.

A non-collocated load cell is often used to measure the cable tension in exosuits \citep{Choi2019, Hosseini2020, Xiloyannis2019b, Georgarakis2022, Wu2018}. A common control framework, referred to as indirect force control \citep{Calanca2016}, closes an outer force loop on the load cell and an inner velocity loop on the motor encoder to render a desired interaction force. Directly measuring the output tension of a Tendon Driver Unit (TDU) allows for disturbances, such as motor inertia and transmission friction, to be compensated for by the controller without a disturbance model. This approach has been demonstrated in many exosuits \citep{Xiloyannis2019b, Georgarakis2022, Wu2018}. Georgarakis and colleagues showed that the tension tracking root-mean-square error (RMSE) can be brought as low as 3.4N (16\%\footnote{This percentage value has been computed based on the original data available to the authors.}) during trajectories that represent activities of daily living (ADLs) patterns \citep{Georgarakis2020}. 

However, force-sensing devices add cost and complexity to the system, require careful tuning, and add distal components. The overall system complexity makes the technology transfer from academia to industry challenging \citep{Tong2018}. Furthermore, the presence of distal components may make the device more cumbersome to wear, lowering its acceptability \citep{Ang2023}. 

\subsection{Friction modelling and compensation}
Frictional losses in Bowden sheath transmission are non-negligible, particularly when the sheath undergoes bending. Additionally, the material properties of the sheath and cable can significantly impact the efficiency of the transmission \citep{Chen2014}. Several models have been proposed to characterize the disturbances inside Bowden transmissions and compensate for them. Many of these are dynamic models based on Coulomb friction and the Capstan formula, often including the elastic behavior of the system \citep{Agrawal2010, Buchanan2021, Wu2019}.  

Since this friction is highly dependent on the cable-sheath bending angle, Jeong et al. explored loop routing as a method to make the frictional disturbance more constant with positional changes of the end-effector relative to the actuator of a 2D manipulator robot \citep{Jeong2015}. Although perhaps counter-intuitive, increasing the magnitude of the disturbance in order to reduce its variability made a feed-forward compensation strategy effective. Nevertheless, loop-routing may not be the best strategy for wearable robots, as extra loops in the transmission might introduce snag points resulting in a less discrete suit design. Moreover, a decrease in the efficiency of the transmission would require a higher magnitude of motor torque negatively impacting motor dimensions and weight, as well as energy consumption and battery life. 

Removing the load cell from cable-driven upper limb rigid exoskeletons has been recently attempted by some research groups \citep{Dezman2022, Zhang2020, Orekhov2020}. For example, Zhang et al. designed an actuation unit used within a 4-DOF arm exoskeleton, which includes a compact rotary series elastic actuator and Bowden cable with a simplified Bowden-sheath friction model in order to estimate the frictional losses in the transmission \citep{Zhang2020}. To the best of our knowledge, no upper limb cable-driven exosuits implement load-cell-free tension control. 

\subsection{Objective}
The objectives of this work are twofold: 

\begin{enumerate}[i) ]
    \item to remove the load cell from the control architecture of a cable-driven upper limb exosuit and to prove the feasibility of the approach;
    \item to assess the efficacy of the novel exosuit and control approach in supporting humeral elevation movements in healthy participants.
\end{enumerate}

This paper presents a design and model-based control approach to remove the load cell from a cable-driven exosuit. A straightforward and reproducible friction model identification procedure was conducted. This was integrated in a model-based tension controller to regulate the cable tension of the exosuit. The feasibility of the approach was proven on healthy participants by assessing the ability of the controller to track the desired tension at different movement speeds and levels of support. The efficacy of the exosuit was also investigated by evaluating muscular activation via surface Electromyography (sEMG) and movement kinematics with and without the suit. Additionally, the comfort and perceived physical exertion were assessed through questionnaires.

\section{Materials and Methods}

\subsection{Exosuit design}
\label{subsec:exo_design}

\begin{figure}
    \begin{subfigure}{0.46\textwidth}
    \centering
        \includegraphics[width=\textwidth]{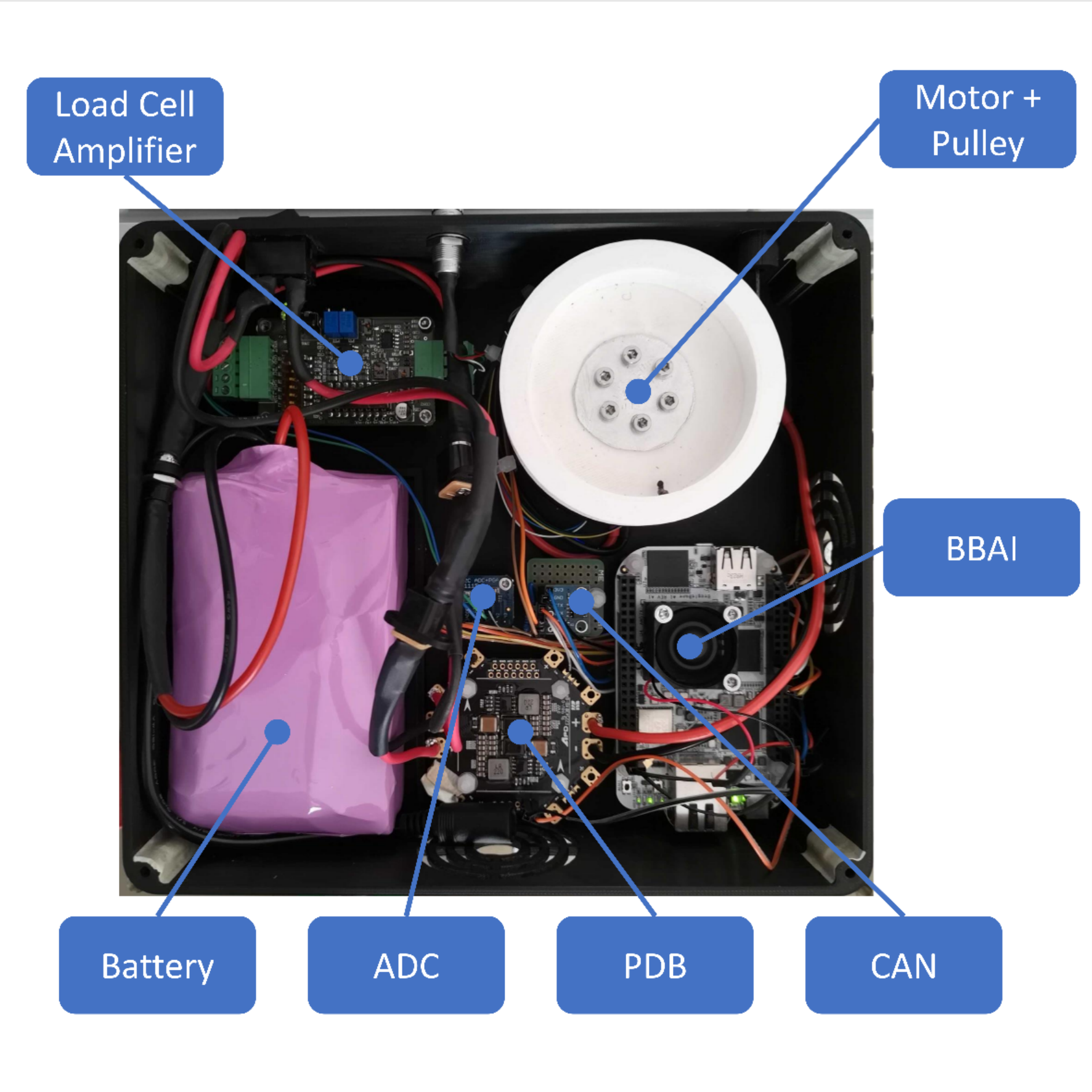}
        \caption{}
        \label{fig:tdu_A}
    \end{subfigure} 
    \begin{subfigure}{0.52\textwidth}
        \centering
        \includegraphics[width=\textwidth]{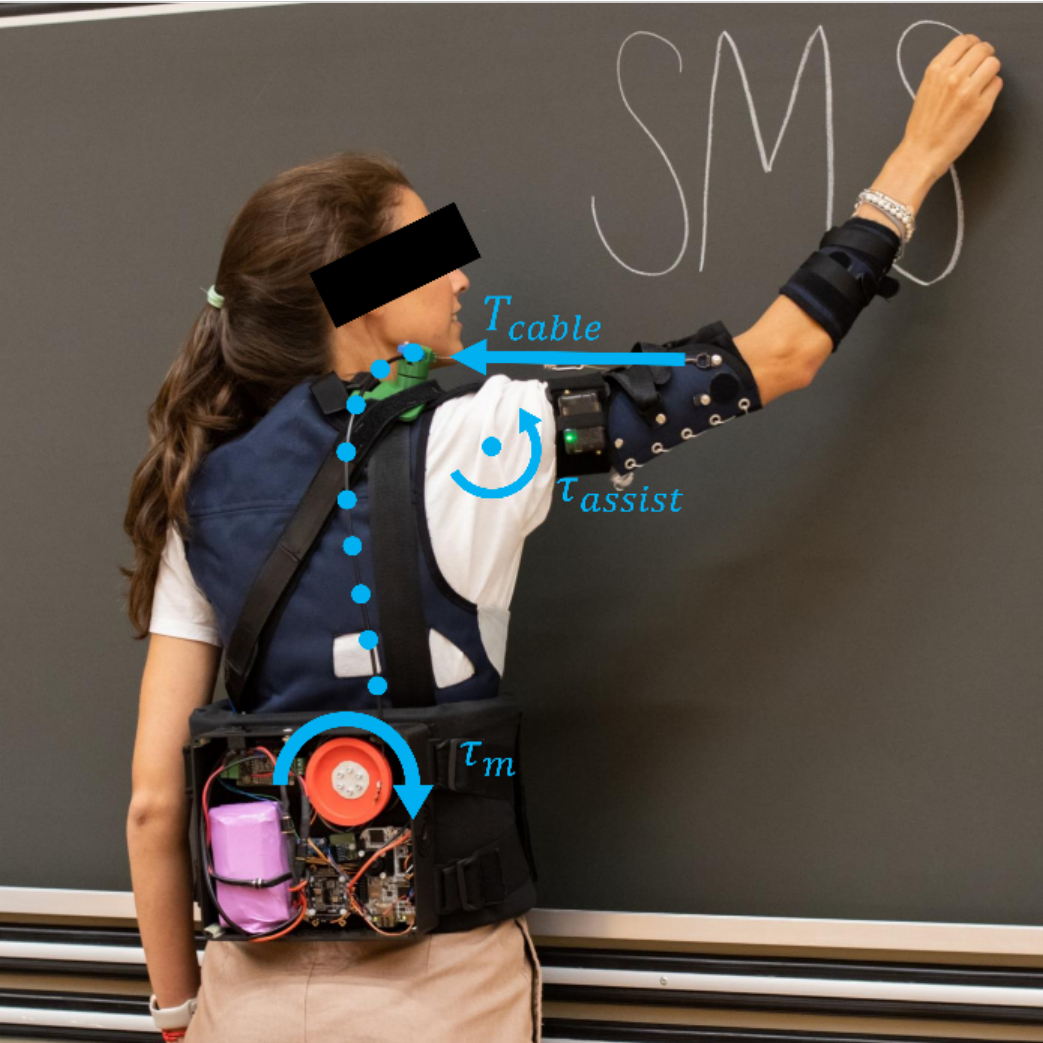}
        \caption{}
        \label{fig:tdu_B}
    \end{subfigure}
    \centering
    \begin{subfigure}{0.39\textwidth}
        \centering
        \includegraphics[width=\textwidth]{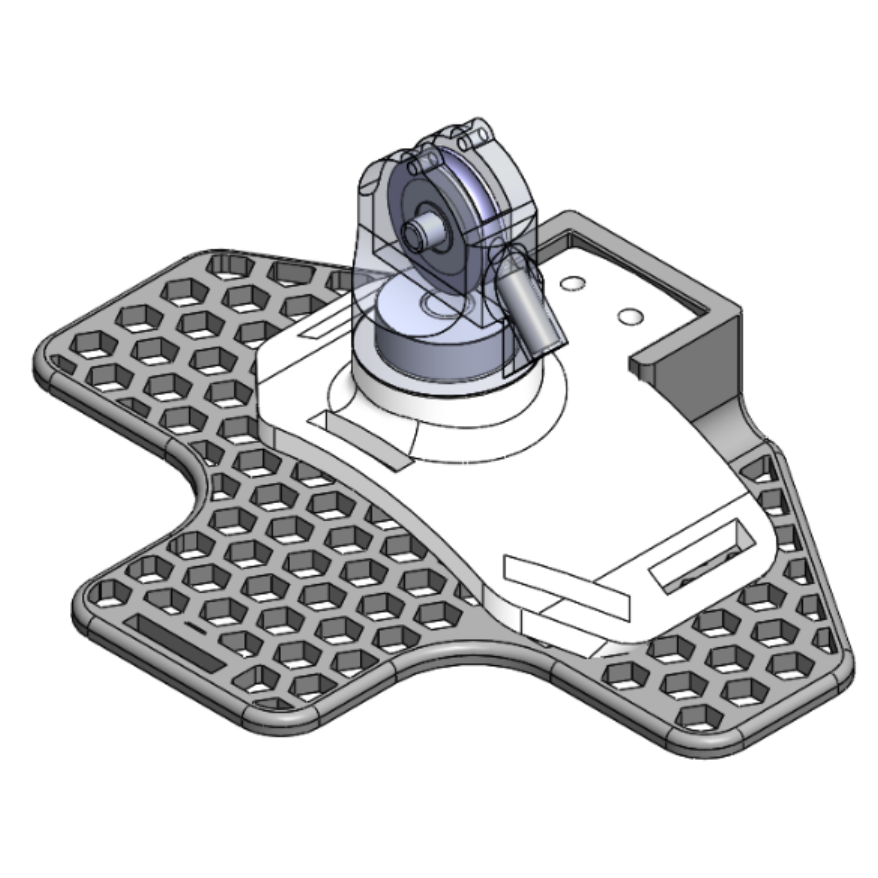}
        \caption{}
        \label{fig:tdu_C}
    \end{subfigure}     
    \begin{subfigure}{0.59\textwidth}
        \centering
        \includegraphics[width=\textwidth]{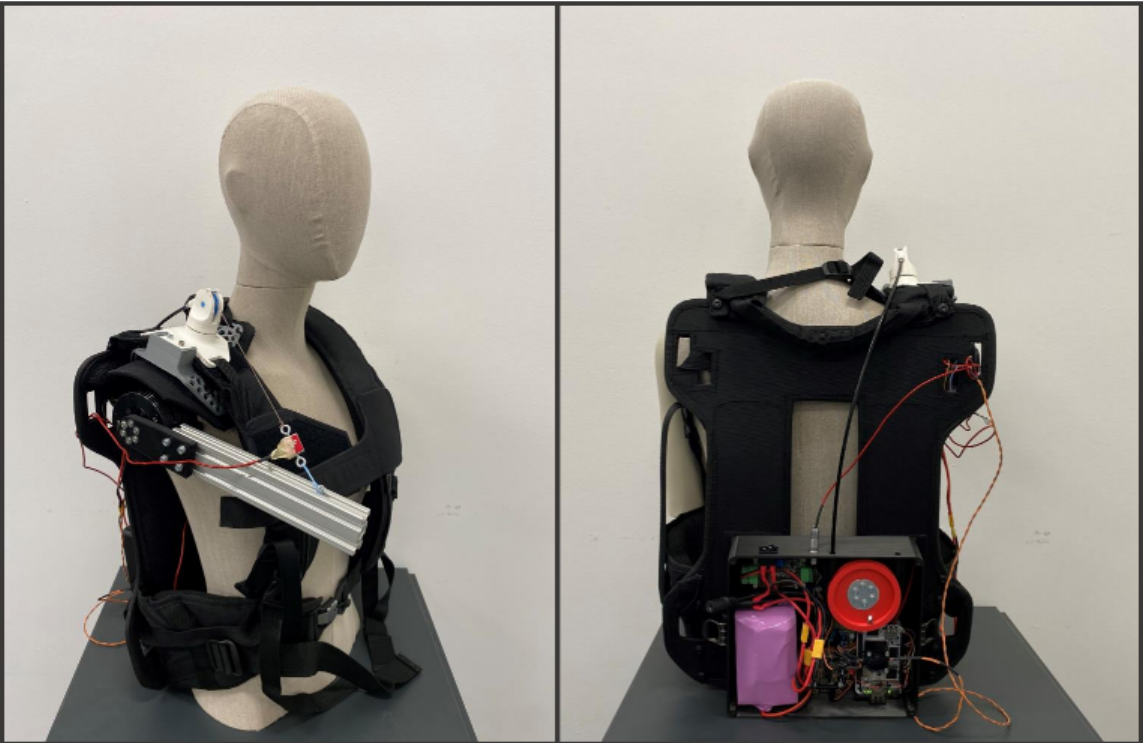}
        \caption{}
        \label{fig:tdu_D}
    \end{subfigure}
    \caption{(\subref{fig:tdu_A}) Picture of the TDU in the final assembled configuration with major electromechanical components labeled. (\subref{fig:tdu_B}) Functioning principle of the exosuit. Cable tension, generated in the TDU by the motor, is transmitted along the body through the Bowden sheath system. This results in an assistive torque about the glenohumeral joint of the wearer, supporting the arm against gravity. (\subref{fig:tdu_C}) Rendering of the swivelling shoulder cuff in Solidworks 2021. (\subref{fig:tdu_D}) Setup for the model identification experiments. The mannequin, modified with a motorized arm, wears the exosuit.}
    \label{fig:tdu_A_B}
\end{figure}

The exosuit designed for this study was composed of:

\begin{enumerate}[i) ]
    \item a TDU (Fig. \ref{fig:tdu_A});
    \item a cable and Bowden sheath used to transmit forces from the TDU into supportive torques on the arm (Fig. \ref{fig:tdu_B});
    \item textile components to secure the TDU and cable to the body of the user.
\end{enumerate}

The TDU actuator was a brushless DC motor (AK60-6 V1.1, T-motor, China) with a 3D-printed poly-lactic acid (PLA) pulley with a diameter of 70mm. This quasi-direct drive (QDD) motor has a reduction of 6:1 and offers a wider control bandwidth and improved torque control with respect to high-gear-ratio brushless DC motors \citep{Yu2020}.

The microcontroller was a BeagleBone AI (BeagleBoard.org, Oakland Twp, MI) using a Controller-Area-Network (CAN) bus (SN65HVD230 3.3V breakout module, Arduino) for motor communications. A 24V, 7Ah lithium-ion battery (Seilylanka) provided power which was distributed with a power distribution board (PDB500, Advanced Power Drives PL, Australia).  The humeral angle of elevation was measured with an IMU strapped to the upper arm (NGIMU, x-io Technologies, UK). A load cell (LSB201, Futek CA, USA), signal amplifier (IAA100, Futek CA, USA), and analog-to-digital converter (16-bit ADS1115, Adafruit Industries NY, USA) were used to measure cable tension. The load cell was not used in the controller but to identify the model parameters and for evaluation purposes.

A backpack frame (B and W International
BPS/5000) and an arm cuff designed by Georgarakis et al. were used as textile components \citep{Georgarakis2022}. The TDU was attached at the lumbar level to keep the system weight closer to the center of mass of the user.

The cable (1.5mm Dyneema cable, ExtremTextil) was guided through a Bowden sheath (Slickwire, SRAM), selected for its high stiffness based on the recommendations from \citep{Chen2014}. The cable exit was secured above the shoulder with a PLA 3D-printed shoulder cuff (Fig. \ref{fig:tdu_C}) designed to minimize frictional losses. The cuff has a bearing to adapt to the humeral plane-of-elevation (defined according to the International Society of Biomechanics \citep{Wu2005} and a pulley at the end of the Bowden sheath to redirect the cable to minimize the bending angle of the sheath.

\subsection{Transmission model and parameter identification}
\subsubsection{Transmission model}
The transmission model maps the desired motor torque ($\tau_{des}$) to the cable tension acting on the arm ($T_{out}$). Inertial effects were assumed negligible since a QDD motor was used. The input tension ($T_{in}$) is related to the desired motor torque and pulley radius ($R_{p}$):
\begin{equation}
    T_{in} = \frac{\tau_{des}}{R_p} 
\end{equation}

The output tension ($T_{out}$) of a Bowden-sheath system is:
\begin{equation}
    \label{eq:model_final}
    T_{out} = T_{in} \left(1 - sgn(V_{cable}) \frac{2 \mu  sin\left(\frac{\phi}{2}\right)}{1 + sgn(V_{cable}) \mu sin\left(\frac{\phi}{2}\right)}\right)
\end{equation}

where $V_{cable}$ is the relative velocity between the cable and sheath, $\mu$ is the Coulomb friction coefficient, and $\phi$ is the sheath total wrapping angle. The full derivation of Eq. (\ref{eq:model_final}) is reported in Section 1 of Supplementary Materials.
The output tension is linearly proportional to the input tension, assuming that $\mu$ and $\phi$ are constant with time, but dependent as well on the direction of cable travel. Ultimately, the inverse model was identified, relating the desired cable output tension and the commanded motor torque.

\subsubsection{Data acquisition}
A motorized human-like mannequin test bench was developed for the identification procedure (Fig. \ref{fig:tdu_D}).

The TDU motor was controlled in trajectory to achieve a set of quasi-constant spool velocities ($\pm$0.5, 1, 2, 3, 4, 5, 6 rad/s) across the range of motion (from 20$^\circ$ to 110$^\circ$) of the driven arm. Each velocity condition was repeated five times. The mannequin motor was controlled in torque to apply a set of constant loads (0.5, 1, 2, 3, 4 Nm) for each velocity condition. 

\subsubsection{Model identification}
\label{sec:model_identification}
The output tension measured by the load cell ($T_{out}$), the estimated torque applied by the TDU motor($\tau_m$), and the velocity estimated from the motor encoder were filtered with a zero-phase third-order low-pass Butterworth filter with a cutoff frequency of 5 Hz. To take into account only data acquired in steady-state conditions, a set of conditions was applied to select relevant data points: 
\begin{itemize}
    \item $TDU_{speed} > 15\frac{^\circ}{s} $,
    \item $TDU_{acc} < 15\frac{^\circ}{s} $,
    \item $20^\circ < \theta_{AOE} < 90^\circ$,
    \item $TDU_{torque} > 0 Nm$, 
\end{itemize}

where $TDU_{speed}$ is the TDU motor speed, $TDU_{acc}$ is the TDU motor acceleration, $\theta_{AOE}$ is the mannequin humeral angle of elevation estimated by the IMU, and $TDU_{torque}$ is the TDU motor torque.
The linear fitting between the desired output cable tension and the commanded motor torque was carried out in MATLAB R2023a (Mathworks inc.) using the \textit{robustfit} function. Two regression lines were estimated, one for raising and one for lowering.

\subsection{Control architecture}
\label{sec:control}
\begin{figure}[h]
    \centering\includegraphics[width=\textwidth]{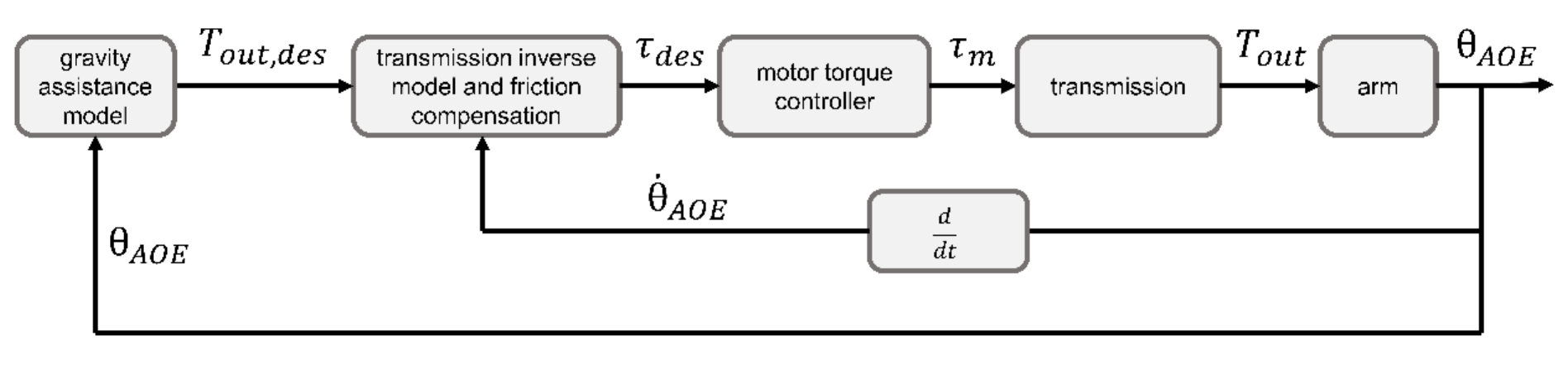}
    \caption{\label{fig:controller}Control block diagram. The "gravity assistance model" block maps the current humeral angle of elevation to the desired output tension based on the anthropometrics of the user and desired support level. The desired tension is sent to the "transmission inverse model and friction compensation block" which computes the desired motor torque as a function of the humeral elevation speed. The desired torque is sent to the motor which runs a low-level torque controller. Finally, the motor output torque acts on the cable through the pulley and is transmitted to the arm through the Bowden sheath mechanism.}
    
\end{figure}

The TDU controller (Fig. \ref{fig:controller}) included:

\begin{enumerate}[i) ]
    \item a gravity assistance block that computes the desired cable tension ($T_{out,des}$) from the humeral angle of elevation ($\theta_{AOE}$);
    \item the transmission inverse model and friction compensation block identified as described in Section \ref{sec:model_identification};
    \item an open-loop torque controller embedded in the motor taking $\tau_{des}$ as input and controlling the motor phase currents.
\end{enumerate}

 The desired motor torque ($\tau_{des}$) depends on whether the arm is being raised or lowered:
 
\begin{equation}
\begin{cases}
    \tau_{des,lowering} = b_{lowering} + m_{lowering} \cdot T_{out,des}, & \text{if $\dot\theta_{AOE} < 0$} \\
    \tau_{des,raising} = b_{raising} + m_{raising} \cdot T_{out,des}, &  \text{if $\dot\theta_{AOE} > 0$}
\end{cases}
\end{equation}

The two regression lines are linked with a sigmoid, ensuring a smooth transition of the commanded torque depending on the arm velocity:

\begin{equation}
    \tau_{des} = \tau_{des,lowering} + \frac{\tau_{des,raising} - \tau_{des,lowering}}{1 + e^{-B \cdot (\dot{\theta}_{AOE} - M)}}
\end{equation}

The shape parameters of the sigmoid coupling function can be set by specifying the humeral elevation velocities at which 1\% and 99\% of the sigmoid extremal values should be achieved ($\dot{\theta}_{AOE,001}$ and $\dot{\theta}_{AOE,099}$ respectively):

\begin{equation}
    M =\frac{\dot{\theta}_{AOE,001} + \dot{\theta}_{AOE,099}}{2}
\end{equation}
\begin{equation}
    B = \frac{\mathrm{ln}(99)}{\dot{\theta}_{AOE,001} - M}
\end{equation}

 For this study $\dot{\theta}_{AOE,001}$ and $\dot{\theta}_{AOE,099}$ were set to -1 $\frac{rad}{s}$ and 1 $\frac{rad}{s}$, respectively. Altering the width of the sigmoid affects how quickly the compensation switches between raising and lowering movements, and shifting the sigmoid affects the amount of compensation while raising, lowering, and in static positions. The torque applied by the motor ($\tau_{m}$) is transmitted to the arm of the user and it results in a tension at the anchor point ($T_{out}$).

\subsection{System evaluation tests on healthy participants}
\subsubsection{Protocol}
\begin{figure}[h]
\centering
    \includegraphics[trim={0cm 2cm 0cm 2cm},clip,width=\textwidth]{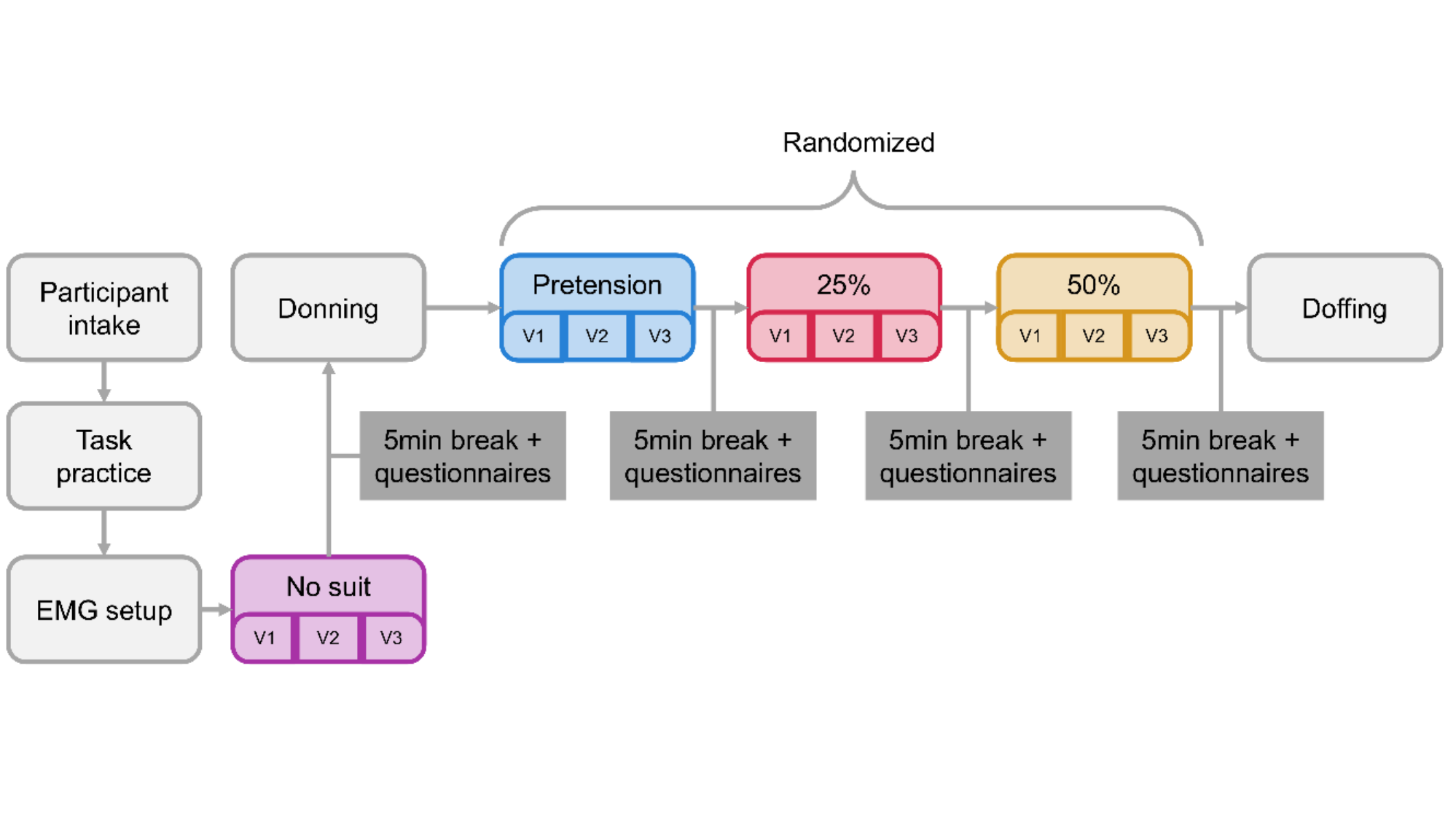}
    \caption{\label{fig:protocol} Study protocol. Participants were first explained the full protocol and had the opportunity to perform a training task without the suit for familiarization. The task was repeated for four conditions: no-suit (no), pretension of 10 N (pre), 25\% gravity support (25\%), 50\% gravity support (50\%). The no-suit condition was always performed as the first to measure the baseline EMG values of each participant. The other three conditions were randomized in order to exclude effects from habituation to the device. A 5-minute break was given to each participant between conditions while a set of questionnaires were administered. For each support condition, three movement speed conditions were tested. The slow condition had a peak speed of 60$\frac{^\circ}{s}$ (V1), the medium condition 120$\frac{^\circ}{s}$ (V2), and the fast condition 180$\frac{^\circ}{s}$ (V3). At the end of the study the exosuit and EMG system was removed, and the participant thanked with a cookie.}
\end{figure}

To evaluate the feasibility of such a controller, experiments were conducted on healthy volunteers. The protocol (Fig. \ref{fig:protocol}) was approved by the ethical committee of Politecnico di Milano (opinion n. 46/2022, 16/11/2022). Each participant read and signed a written informed consent.

The task consisted of standing upright and tracking a forward humeral elevation reference trajectory shown on a screen while holding a 0.5 kg weight. The actual humeral angle of elevation estimated by the IMU sensor was provided as feedback on the same screen. The minimum-jerk trajectory transitioned between 20$^\circ$ and 100$^\circ$ arm elevation, and was repeated ten times for each peaks speed and support level. The participants were verbally instructed to keep the elbow fully extended, but relaxed.

The EMG electrodes were applied to the muscles involved in humeral elevation: agonist muscles (anterior deltoid, medial deltoid, trapezius, pectoralis major), antagonist muscles (posterior deltoid, latissimus dorsi), and elbow stabilizers (biceps, triceps), following the international SENIAM recommendations \citep{Hermens1999}. The EMG acquisition system was a SAGA 32+/64+ (TMS International, Netherlands) sampling the signal at a frequency of 2000 Hz and synchronized with the TDU.

\subsubsection{Data processing}
EMG data were pre-processed in MATLAB R2023a (Mathworks inc.) to extract the EMG envelopes of the selected muscles. A third-order zero-phase Butterworth bandpass filter (20-400Hz) was applied to the raw data which was then rectified and smoothened with a sliding 100ms-wide RMS window \citep{Burden2017}. Additionally, for the pectoralis major and the latissimus dorsi, the ECG artifact was removed by means of adaptive template subtraction with the Cardiac artifact removal toolbox \citep{Petersen2020, githubGitHubImeluebeckecgremoval}.
The first of ten repetitions for each condition-velocity set was removed from the analysis because of adaptation effects happening at the beginning of the exercise when condition and/or speed were changed \citep{Emken2007}. 

The EMG envelope amplitude was normalized to the average maximum peak of the no-suit condition, separately for each participant and velocity. 

\subsubsection{Outcome measures}
The following outcome measures were computed for each repetition: 
\begin{itemize}
    \item RMSE between the desired and the measured tension and torque to evaluate the controller tracking performance split in the raising and lowering phases;
    \item RMSE between the desired and the measured humeral angle of elevation and SPARC index \citep{Balasubramanian2015} computed from the speed of humeral elevation estimated from the IMU ($\dot\theta_{AOE}$) to evaluate the effect of the suit on the kinematics of the arm;
    \item average integral of the sEMG (iEMG) on a normalized time vector to evaluate the effect of the suit on the activation of the selected muscles split in the raising and lowering phases;
    \item perceived discomfort (modified Nordic questionnaire from \citep{KUORINKA1987233}), perceived physical exertion (Borg Rate of Perceived Exertion, or RPE from \citep{borg1982psychophysical}), and perceived muscular fatigue to evaluate the perception of the user. Details on the questionnaires can be found in Section 2 of Supplementary Materials. 
\end{itemize}

\subsubsection{Statistical analysis}
Generalized linear mixed models (GLMM) were defined in SPSS Statistics 28 (IBM) to evaluate whether the iEMG (split into raising, lowering, and entire movement cycle) of each muscle, the humeral elevation angle RMSE, and the SPARC-index were significantly different across conditions and speeds without the assumption of normality, taking into account repeated measures and multiple effects. In the analysis, the iEMG, the RMSE, and the SPARC-index were considered as the target for each model following a gamma regression distribution with a log link to the linear model. Condition, speed, and the interaction of the two were considered fixed effects while the participants were considered as random effects by associating their ID to the intercept with a variance component covariance type. A pairwise comparison with the sequential Bonferroni correction for multiple comparisons was conducted on the effects that showed a p-value lower than 0.05.

The questionnaire results were compared across conditions by means of a Friedman test in MATLAB R2023a (Mathworks Inc.). If the p-value was lower than 0.05, a pairwise comparison was conducted by means of the \textit{multcompare} function applying the Bonferroni correction for multiple comparisons.

\section{Results}
\subsection{Model identification on mannequin}
\begin{figure}[h]
    \centering
    \includegraphics[width=0.8\textwidth] {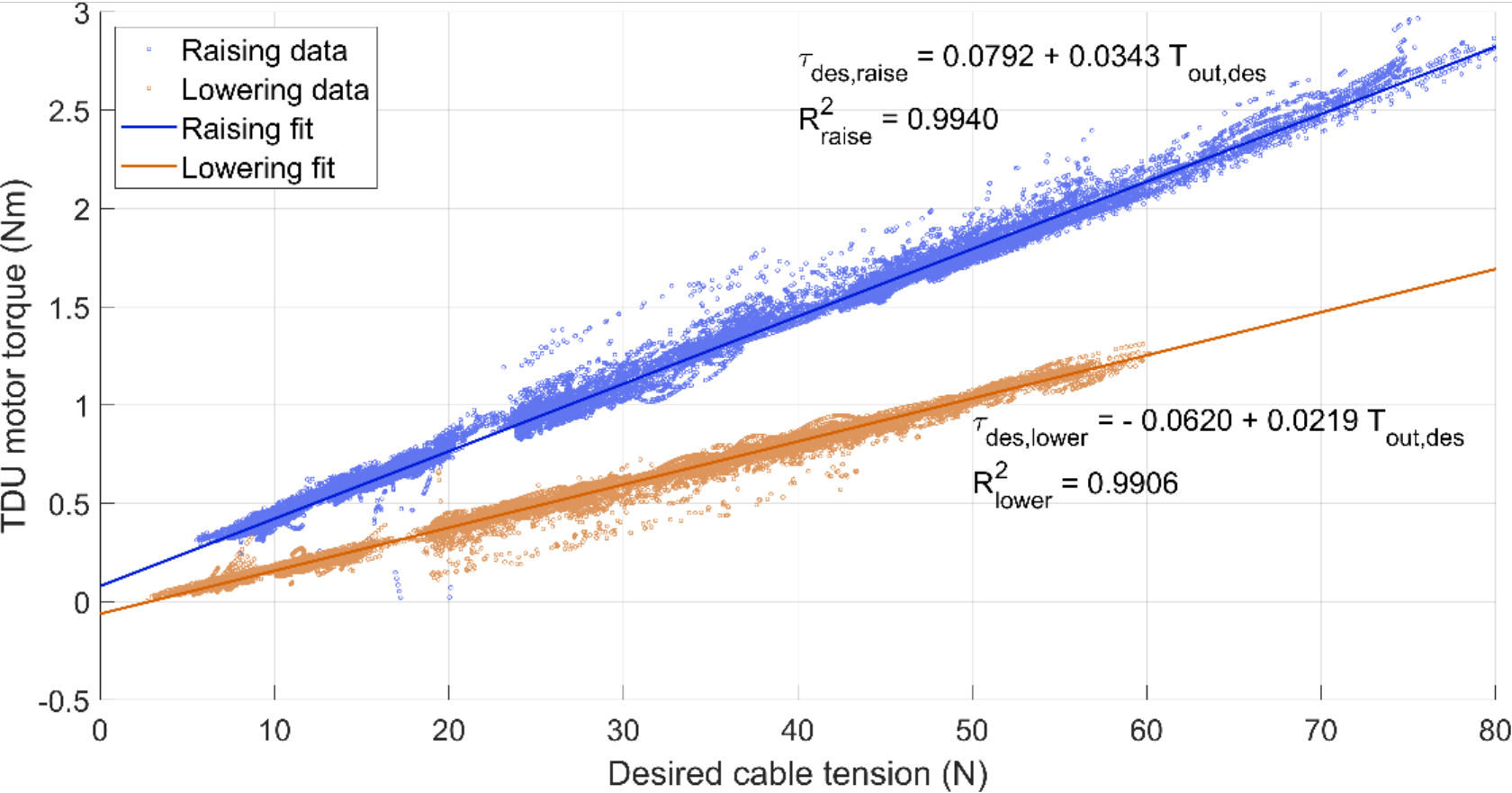}
    \caption{Linear regressions of the desired cable tension ($T_{out,des}$) and the TDU motor torque ($\tau_{des}$) required. The graph axes are presented this way to reflect the controller architecture (Fig. \ref{fig:controller}). The data and regression line for the raising phase is shown in blue, while the lowering phase is shown in orange. Slope and intercept are displayed along with goodness of fit ($R^2$).}
    \label{fig:linear_regression}
\end{figure}

The identified regression lines for raising and lowering had a goodness of fit of 0.9940 and 0.9906, respectively (Fig. \ref{fig:linear_regression}). The slopes and the intercepts of the two regression lines were used in the control strategy described in Section \ref{sec:control}.

\subsection{System evaluation experiments on healthy participants}
Eighteen healthy volunteers (12 F / 6 M) were recruited for the study. The experiment was interrupted for one participant due to electrode detachment and discontinued. The participants' mean age was 29.33 years (SD = 7.04), the mean height was 1.70m (SD = 0.11), and the mean weight was 63.23 kg (SD = 11.51). 

\begin{figure} 
    \centering
    \includegraphics[width=\textwidth]{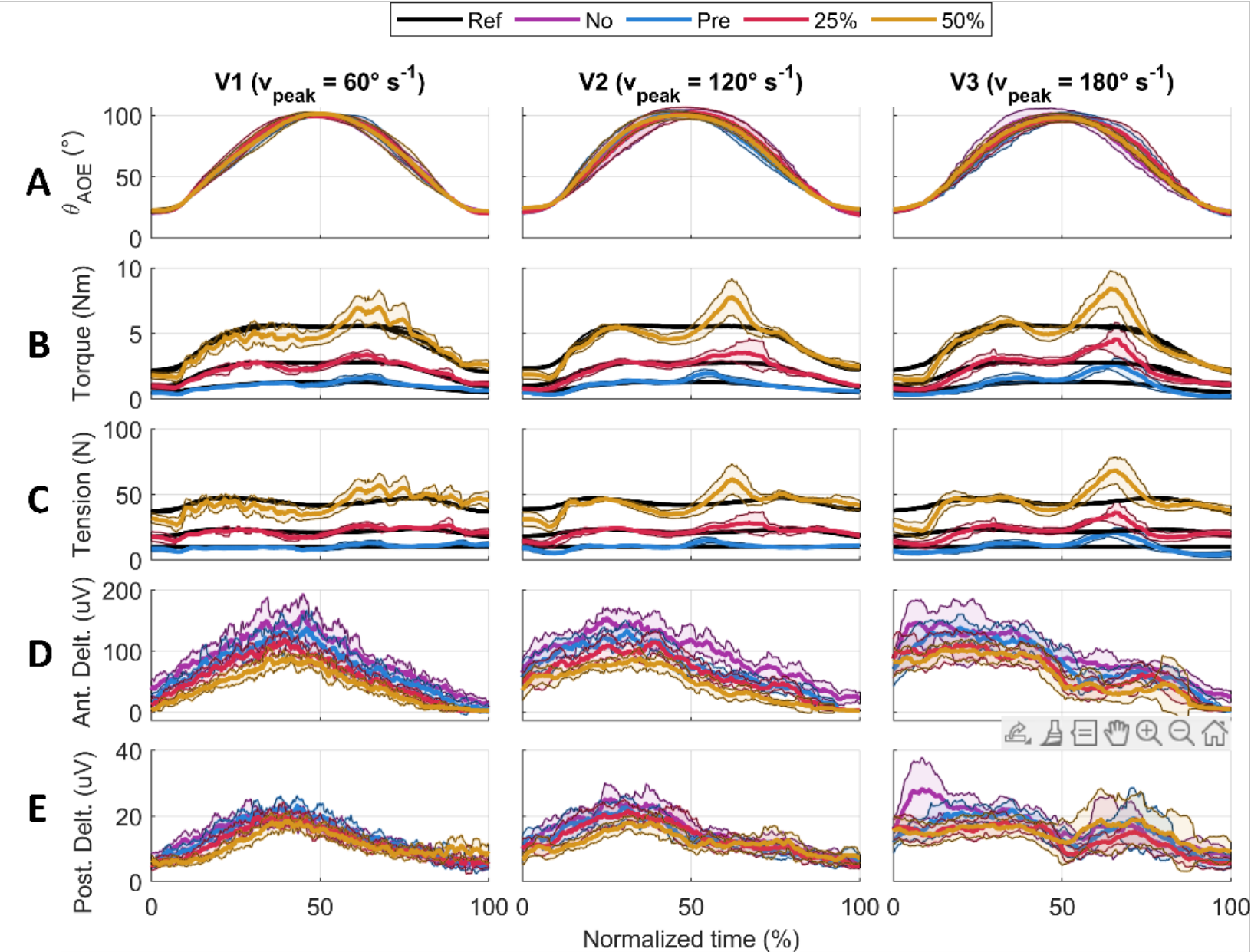}
    \caption{Results for one participant averaged over the nine repetitions. The solid lines represent the average value, the shaded area is the standard deviation. The four support conditions, "No", "Pre", "25\%", and "50\%", are represented with the colors purple, blue, red, and yellow, respectively. Each column shows a velocity condition (V1, V2, V3). \textbf{A)} Humeral angle of elevation ($\theta_{AOE}$). \textbf{B and C)} Supporting torque acting at the shoulder joint and the cable tension acting on the anchor point, respectively, where the black line indicates the reference signal. \textbf{D and E)} EMG envelope of the anterior deltoid and the posterior deltoid, respectively.}
    \label{fig:single_part_results}
\end{figure}

For each participant, the humeral angle of elevation, the supportive torque, the cable tension, and the muscle sEMG were measured. The data was averaged over the nine repetitions and resampled onto a normalized time vector for each participant, condition, and speed (Fig. \ref{fig:single_part_results}).

\subsubsection{Assistive tension and torque tracking}
The shoulder torque RMSE and cable tension RMSE increased with support and velocity for the entire movement cycle. The overall torque RMSE ranged from a minimum of 0.29Nm to 0.90Nm, while the cable tension RMSE ranged from a minimum of 2.75 N to 9.07 N (Table \ref{tab:torque_tension_tracking_RMSE_combined}). If considered as percentage errors compared to the reference signal, these errors were 27.5\% and 23.4\%, respectively (see Supplementary Materials Table S1). 

When considering the movement split into the raising and lowering phases, the torque and tension RMSE values were always higher when lowering. Peaks in the applied torque and tension exceeding the reference were observed around the onset of lowering at approximately 60\% of the normalized time of the entire movement (Fig. \ref{fig:single_part_results} panels B and C, see Supplementary Materials Tables S2 and S3 for computed average values).

\begin{table}[htbp]
\centering
\begin{tabular}{@{}cccccccccc@{}}
\toprule
\multicolumn{1}{l}{} & \multicolumn{3}{c}{\makecell{\textbf{V1} \\ ($v_{peak}$ = 60$\frac{^\circ}{s}$)}}           & \multicolumn{3}{c}{\makecell{\textbf{V2} \\ ($v_{peak}$ = 120$\frac{^\circ}{s}$)}}           & \multicolumn{3}{c}{\makecell{\textbf{V3} \\ ($v_{peak}$ = 180$\frac{^\circ}{s}$)}}           \\ \midrule
\makecell{\textbf{Torque} \\ \textbf{RMSE} \textbf{(Nm)} } & \textbf{Entire} & \textbf{Raise} & \textbf{Lower} & \textbf{Entire} & \textbf{Raise} & \textbf{Lower} & \textbf{Entire} & \textbf{Raise} & \textbf{Lower} \\ 
\textbf{pre} & 0.29 & 0.23 & 0.32 & 0.40 & 0.25 & 0.47 & 0.69 & 0.43 & 0.84 \\
\textbf{25\%} & 0.42 & 0.38 & 0.44 & 0.58 & 0.41 & 0.68 & 0.74 & 0.50 & 0.91 \\ 
\textbf{50\%} & 0.66 & 0.64 & 0.66 & 0.71 & 0.54 & 0.82 & 0.90 & 0.57 & 1.11 \\ \midrule
\makecell{\textbf{Tension} \\ \textbf{RMSE} \textbf{(N)} } & \textbf{Entire} & \textbf{Raise} & \textbf{Lower} & \textbf{Entire} & \textbf{Raise} & \textbf{Lower} & \textbf{Entire} & \textbf{Raise} & \textbf{Lower} \\
\textbf{pre} & 2.75 & 2.37 & 2.98 & 3.58 & 2.90 & 3.92 & 6.15 & 4.51 & 7.20 \\
\textbf{25\%} & 4.09 & 3.83 & 4.14 & 5.31 & 4.70 & 5.61 & 6.74 & 5.70 & 7.47 \\ 
\textbf{50\%} & 6.14 & 6.03 & 6.10 & 6.46 & 5.94 & 6.72 & 8.17 & 6.80 & 9.07 \\ \bottomrule
\end{tabular}
\caption{\label{tab:torque_tension_tracking_RMSE_combined} Torque and tension tracking RMSE presented for all combinations of velocities and support conditions, averaged across all participants. The results are further split showing the average RMSE of the entire arm cycle, the raising portion, and the lowering portion.}
\end{table}

\subsubsection{Trajectory tracking and smoothness}
\begin{figure}[htbp]
    \centering
    \includegraphics [width=\textwidth]{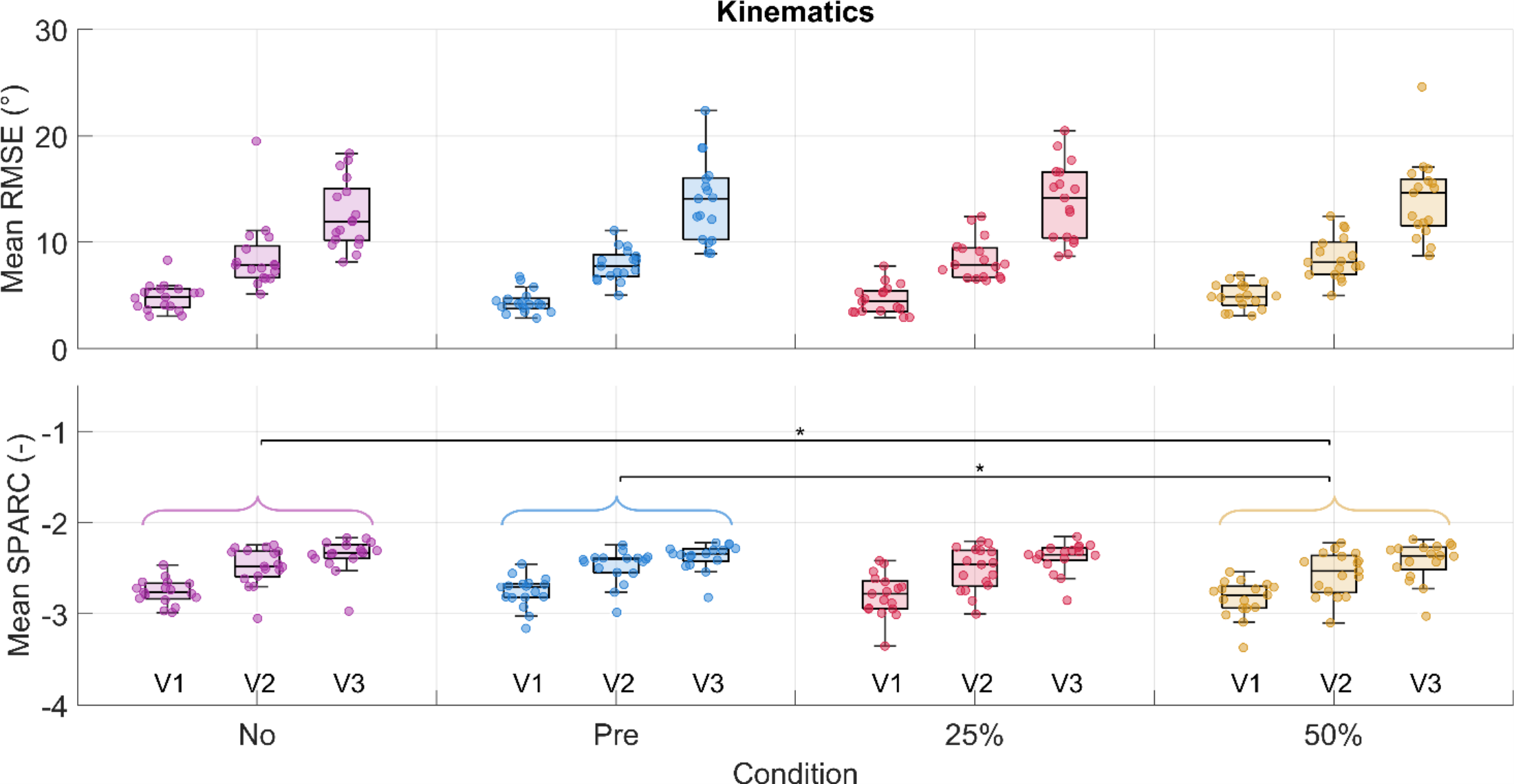}
    \caption{RMSE and SPARC for the humeral angle of elevation are shown in the first and second rows, respectively. Each dot in each boxplot represents one participant. The results are shown subdivided by velocity but the statistical significance level for the pairwise comparison among conditions is shown for the velocities grouped. The following significance codes are used to represent the according ranges of p-values for the post-hoc tests: ** = [0,0.001), * = [0.001, 0.05).}
    \label{fig:kinematics_results}
\end{figure}

A significant difference in the condition effect was found for the SPARC index between the no-suit and the 50\% gravity support conditions, and the pretension and the 50\% gravity support conditions (Fig. \ref{fig:kinematics_results}). SPARC decreased significantly by 3\% between the no-suit and the 50\% gravity support conditions. As for the velocity effect, a significant difference was found both for the RMSE and the SPARC index. No difference was found in the interaction of condition and velocity for both outcomes (detailed results are reported in Supplementary Materials Tables S4, S5, and S6).

\subsubsection{EMG}
Significant differences in the condition and velocity effects were found across the muscles, both in the raising and lowering phases. No differences were found for the interaction effect of condition and velocity (see Supplementary Materials Tables S7, S8, and S9). 

\begin{figure}
    \centering
    \includegraphics[width=\textwidth]{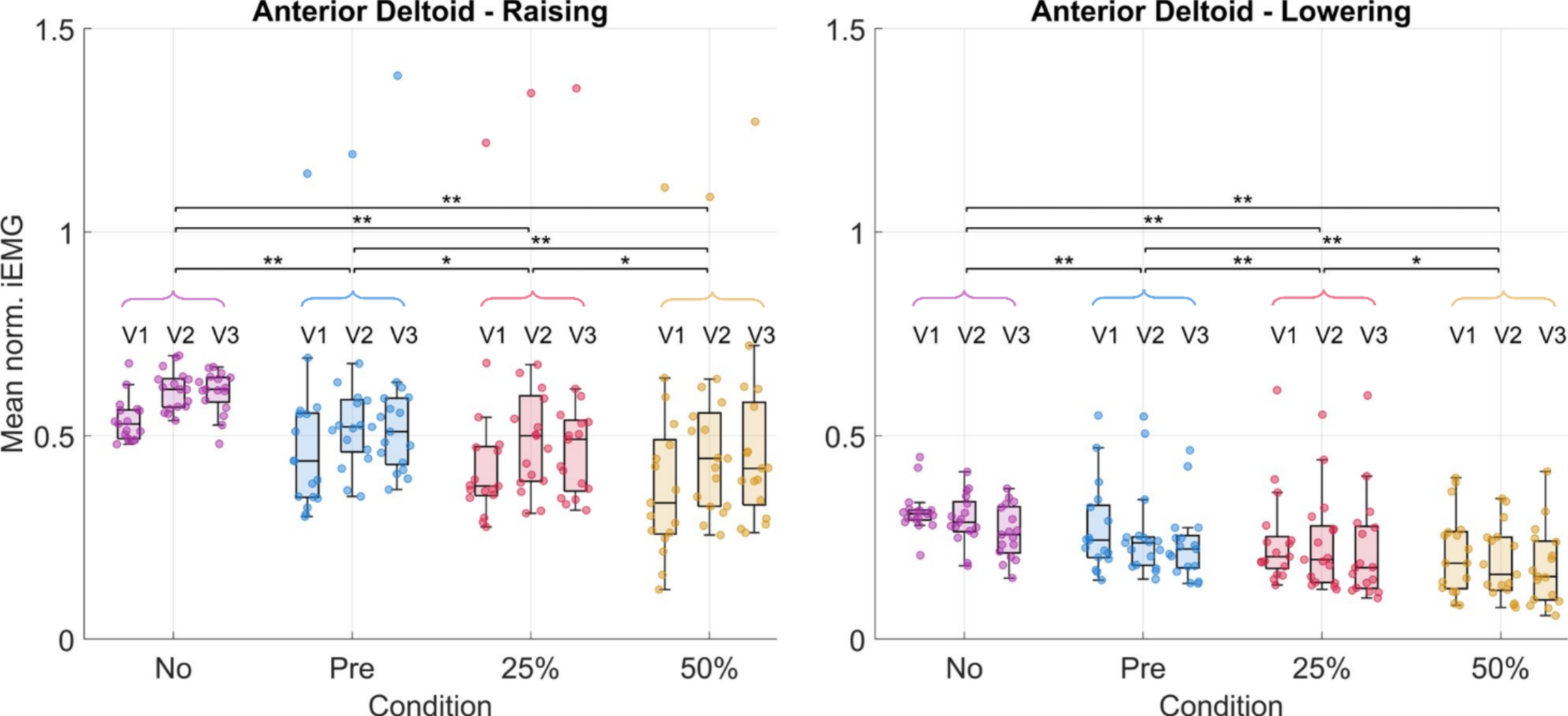}
    \caption{Results in terms of mean normalized iEMG for the anterior deltoid. Each dot in each boxplot represents one participant. The results are shown subdivided by velocity but the statistical significance level for the pairwise comparisons between conditions are shown for the velocities grouped. The following significance codes are used to represent the according ranges of p-values for the post-hoc tests: ** = [0,0.001), * = [0.001, 0.05).}
    \label{fig:ant_delt_iemg}
\end{figure}

Significant decreases in the iEMG from the no-suit condition to the various support conditions were detected for the anterior deltoid (Fig. \ref{fig:ant_delt_iemg}), trapezius, and pectoralis major during the raising and lowering phases along with the entire movement cycle. The greatest changes happened between the no-suit and 50\% support conditions. The anterior deltoid, trapezius, and pectoralis major activities decreased by 30\%, 38\%, and 38\% during raising, and 39\%, 42\%, and 35\% during lowering, respectively. 

\begin{figure}
    \centering
    \includegraphics[width=\textwidth]{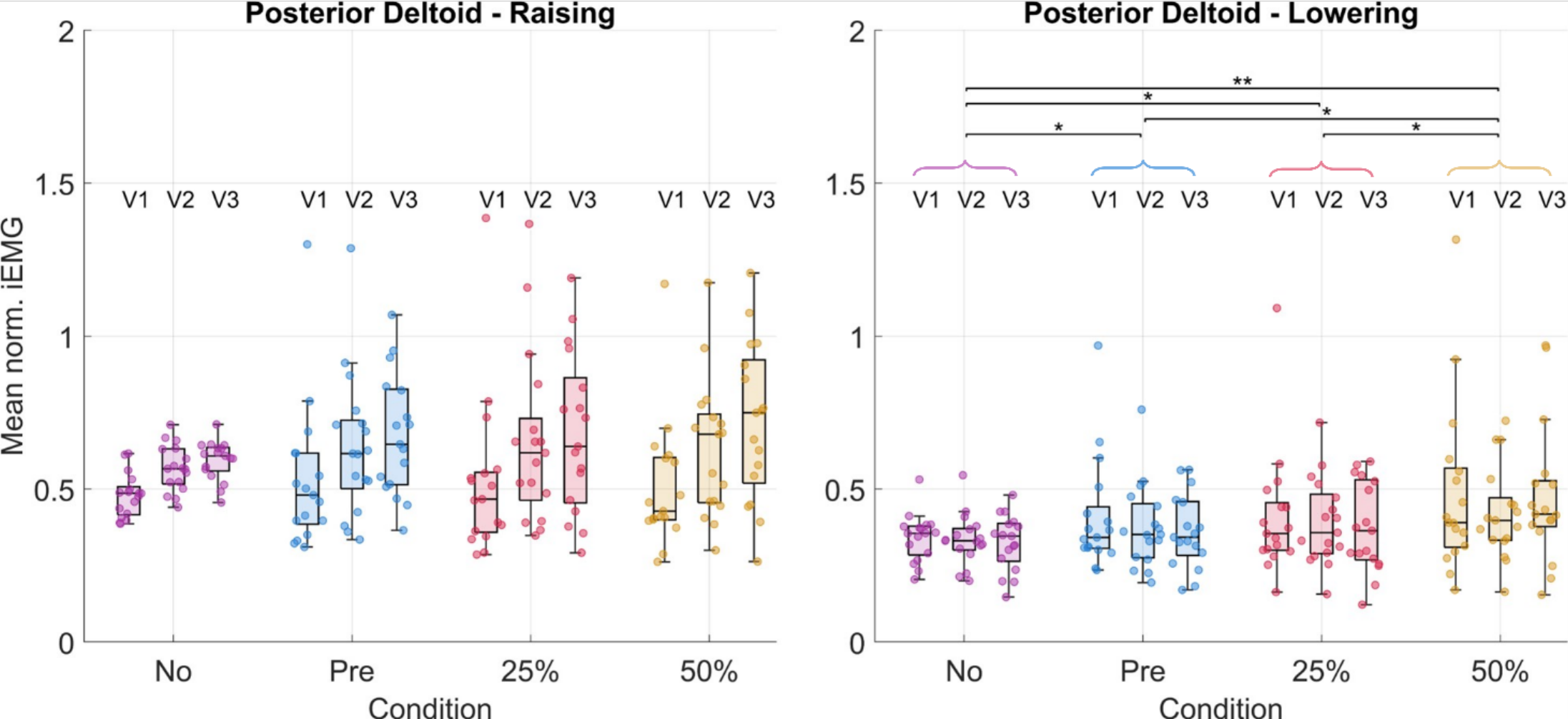}
    \caption{Results in terms of mean normalized iEMG for the posterior deltoid. Each dot in each boxplot represents one participant. The results are shown subdivided by velocity but the statistical significance levels for the pairwise comparisons between conditions are shown for the velocities grouped. The following significance codes are used to represent the according ranges of p-values for the post-hoc tests: ** = [0,0.001), * = [0.001, 0.05).}
    \label{fig:post_delt_iemg}
\end{figure}

The posterior deltoid activity increased significantly by 32\% during lowering (Fig. \ref{fig:post_delt_iemg}). Other muscles, such as the medial deltoid and latissimus dorsi showed some significant differences, though the trends across all post-hoc tests were less clear (changes visualized for all muscles in Tables \ref{tab:iEMG_deltas_visual_raising} and \ref{tab:iEMG_deltas_visual_lowering}, see Supplementary Materials Tables S10, S11, and S12 for numerical values).

\newcommand\Tstrut{\rule{0pt}{2.6ex}}         % = `top' strut
\newcommand\Bstrut{\rule[-0.9ex]{0pt}{0pt}}   % = `bottom' strut

\begin{table}[htbp]
\begin{minipage}[]{0.80\textwidth}
\resizebox{\textwidth}{!}{%
\begin{tabular}{|c|cccccc|}
\multicolumn{7}{c}{\textbf{Raising phase}} \\
\hline
Muscle group & pre-no & 25\%-no & 50\%-no & 25\%-pre & 50\%-pre & 50\%-25\%\Tstrut\Bstrut \\
\hline
Ant. delt.  
& \cellcolor[RGB]{161,255,161}\textbf{**} 
& \cellcolor[RGB]{122,255,122}\textbf{**} 
& \cellcolor[RGB]{76,255,76}\textbf{**} 
& \cellcolor[RGB]{209,255,209}\textbf{*} 
& \cellcolor[RGB]{155,255,155}\textbf{**} 
& \cellcolor[RGB]{197,255,197}\textbf{*} \\

Med. delt.  
& \cellcolor[RGB]{255,223,223} 
& \cellcolor[RGB]{255,243,243} 
& \cellcolor[RGB]{193,255,193} 
& \cellcolor[RGB]{237,255,237} 
& \cellcolor[RGB]{165,255,165}\textbf{**} 
& \cellcolor[RGB]{181,255,181}\textbf{*} \\

Post. delt. 
& \cellcolor[RGB]{255,225,225} 
& \cellcolor[RGB]{255,233,233} 
& \cellcolor[RGB]{255,227,227} 
& \cellcolor[RGB]{249,255,249} 
& \cellcolor[RGB]{255,255,255} 
& \cellcolor[RGB]{255,251,251} \\

Biceps
& \cellcolor[RGB]{255,189,189}\textbf{*}
& \cellcolor[RGB]{255,173,173}\textbf{*}
& \cellcolor[RGB]{255,233,233} 
& \cellcolor[RGB]{255,241,241} 
& \cellcolor[RGB]{215,255,215} 
& \cellcolor[RGB]{203,255,203} \\

Triceps
& \cellcolor[RGB]{241,255,241}
& \cellcolor[RGB]{221,255,221} 
& \cellcolor[RGB]{171,255,171} 
& \cellcolor[RGB]{233,255,233} 
& \cellcolor[RGB]{185,255,185}\textbf{*} 
& \cellcolor[RGB]{205,255,205} \\

Trapezius 
& \cellcolor[RGB]{165,255,165}\textbf{**} 
& \cellcolor[RGB]{131,255,131}\textbf{**} 
& \cellcolor[RGB]{28,255,28}\textbf{**} 
& \cellcolor[RGB]{215,255,215}\textbf{*} 
& \cellcolor[RGB]{94,255,94}\textbf{**} 
& \cellcolor[RGB]{126,255,126}\textbf{**} \\

Lat. dorsi
& \cellcolor[RGB]{239,255,239} 
& \cellcolor[RGB]{253,255,253} 
& \cellcolor[RGB]{165,255,165}\textbf{*} 
& \cellcolor[RGB]{255,243,243} 
& \cellcolor[RGB]{181,255,181}\textbf{*} 
& \cellcolor[RGB]{169,255,169}\textbf{*} \\

Pectoralis maj.
& \cellcolor[RGB]{131,255,131}\textbf{**} 
& \cellcolor[RGB]{100,255,100}\textbf{**} 
& \cellcolor[RGB]{26,255,26}\textbf{**} 
& \cellcolor[RGB]{221,255,221} 
& \cellcolor[RGB]{126,255,126}\textbf{**} 
& \cellcolor[RGB]{157,255,157}\textbf{**} \\

\hline
\end{tabular}%
}
\end{minipage}
\begin{minipage}[]{0.1\textwidth}
\includegraphics[height=3.5cm]{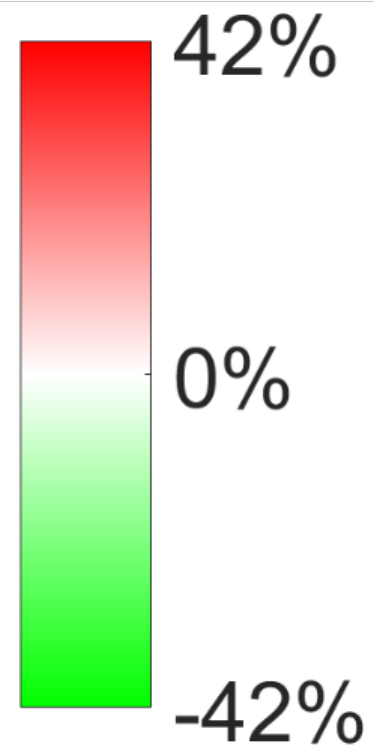}
\end{minipage}
\caption{\label{tab:iEMG_deltas_visual_raising}EMG statistical results for the pairwise post-hoc tests comparing changes in muscular activity across all muscle groups for the raising portion of the arm cycle. The following significance codes are used for the p-values: \textbf{**} = [0, 0.001), \textbf{*} = [0.001, 0.05), and a blank space represents [0.05, 1]. The percentage changes in normalized iEMG, relative to the second condition being compared, is coded in color, with green representing a decrease in iEMG from the second to the first condition, and red representing an increase.}
\end{table}

\begin{table}[htbp]
\begin{minipage}[]{0.80\textwidth}
\resizebox{\textwidth}{!}{%
\begin{tabular}{|c|cccccc|}
\multicolumn{7}{c}{\textbf{Lowering phase}} \\
\hline
Muscle group & pre-no & 25\%-no & 50\%-no & 25\%-pre & 50\%-pre & 50\%-25\%\Tstrut\Bstrut \\
\hline
Ant. delt.
& \cellcolor[RGB]{149,255,149}\textbf{**} 
& \cellcolor[RGB]{90,255,90}\textbf{**} 
& \cellcolor[RGB]{18,255,18}\textbf{**} 
& \cellcolor[RGB]{185,255,185}\textbf{**} 
& \cellcolor[RGB]{98,255,98}\textbf{**} 
& \cellcolor[RGB]{159,255,159}\textbf{**} \\

Med. delt.
& \cellcolor[RGB]{255,189,189} 
& \cellcolor[RGB]{255,233,233} 
& \cellcolor[RGB]{191,255,191} 
& \cellcolor[RGB]{217,255,217} 
& \cellcolor[RGB]{137,255,137}\textbf{**} 
& \cellcolor[RGB]{171,255,171}\textbf{*} \\

Post. delt. 
& \cellcolor[RGB]{255,197,197}\textbf{*}
& \cellcolor[RGB]{255,187,187}\textbf{*} 
& \cellcolor[RGB]{255,62,62}\textbf{**} 
& \cellcolor[RGB]{255,247,247} 
& \cellcolor[RGB]{255,135,135}\textbf{*} 
& \cellcolor[RGB]{255,145,145}\textbf{*} \\

Biceps
& \cellcolor[RGB]{255,157,157}\textbf{*} 
& \cellcolor[RGB]{255,171,171}\textbf{*} 
& \cellcolor[RGB]{255,235,235} 
& \cellcolor[RGB]{245,255,245} 
& \cellcolor[RGB]{189,255,189}\textbf{*} 
& \cellcolor[RGB]{201,255,201} \\

Triceps
& \cellcolor[RGB]{229,255,229} 
& \cellcolor[RGB]{209,255,209} 
& \cellcolor[RGB]{203,255,203} 
& \cellcolor[RGB]{233,255,233} 
& \cellcolor[RGB]{227,255,227} 
& \cellcolor[RGB]{249,255,249} \\

Trapezius
& \cellcolor[RGB]{72,255,72}\textbf{**} 
& \cellcolor[RGB]{40,255,40}\textbf{**} 
& \cellcolor[RGB]{0,255,0}\textbf{**} 
& \cellcolor[RGB]{207,255,207} 
& \cellcolor[RGB]{153,255,153}\textbf{*}
& \cellcolor[RGB]{195,255,195} \\

Lat. dorsi
& \cellcolor[RGB]{255,223,223} 
& \cellcolor[RGB]{255,217,217} 
& \cellcolor[RGB]{255,167,167} 
& \cellcolor[RGB]{255,251,251} 
& \cellcolor[RGB]{255,203,203} 
& \cellcolor[RGB]{255,209,209} \\

Pectoralis maj.
& \cellcolor[RGB]{116,255,116}\textbf{*}
& \cellcolor[RGB]{84,255,84}\textbf{*}
& \cellcolor[RGB]{44,255,44}\textbf{**} 
& \cellcolor[RGB]{215,255,215}
& \cellcolor[RGB]{163,255,163}\textbf{*}
& \cellcolor[RGB]{201,255,201} \\

\hline
\end{tabular}%
}
\end{minipage}
\begin{minipage}[]{0.1\textwidth}
\includegraphics[height=3.5cm]{BardiTableColorbar.pdf}
\end{minipage}
\caption{\label{tab:iEMG_deltas_visual_lowering}EMG statistical results for the pairwise post-hoc tests comparing changes in muscular activity across all muscle groups for the lowering portion of the arm cycle. The following significance codes are used for the p-values: \textbf{**} = [0, 0.001), \textbf{*} = [0.001, 0.05), and a blank space represents [0.05, 1]. The percentage changes in normalized iEMG, relative to the second condition being compared, is coded in color, with green representing a decrease in iEMG from the second to the first condition, and red representing an increase.}
\end{table}

\subsubsection{Questionnaires}
\begin{figure}
    \centering
    \includegraphics[width=0.8\textwidth]{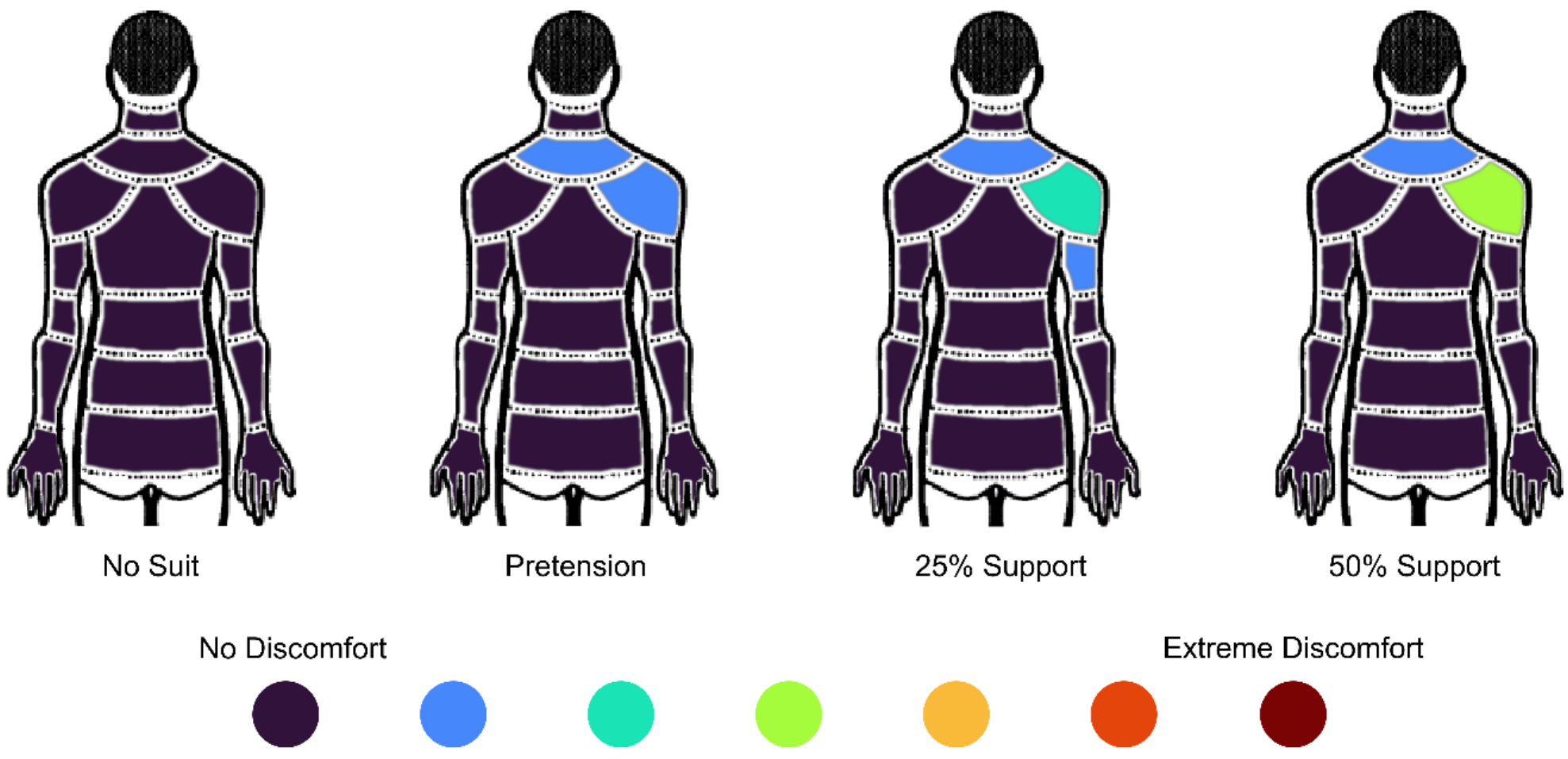}
    \caption{Modified Nordic questionnaire results reporting median score across all participants. The figure is viewed from behind, and the right arm was supported by the exosuit.}
    \label{fig:nordic_results}
\end{figure}

Significant changes in discomfort in the lower neck ($p = 0.001$) and right shoulder ($p < 0.001$)  were highlighted by the statistical analysis of the Nordic questionnaire outcomes  (Fig. \ref{fig:nordic_results}). In particular, the discomfort in the lower neck area significantly increased from the no-suit condition to the pretension ($p = 0.027$), 25\% ($p = 0.002$), and 50\% ($p = 0.006$) support conditions. The discomfort in the shoulder area significantly increased from the no-suit condition to the 25\% ($p = 0.002$), and 50\% ($p < 0.001$) support conditions and also from the pretension condition to the 50\% support condition ($p = 0.007$).

\begin{figure}
    \centering
    \includegraphics[width=0.8\textwidth]{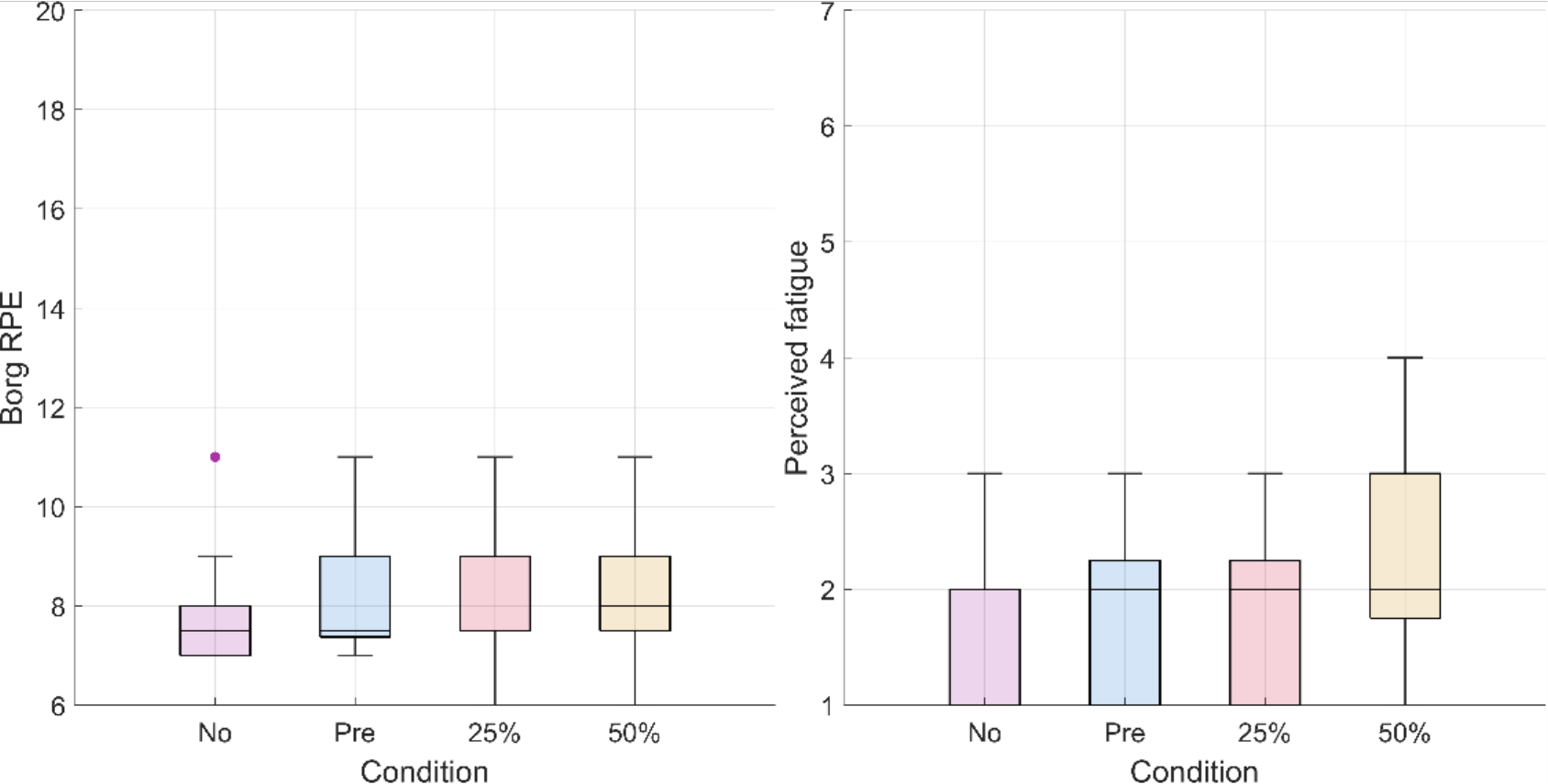}
    \caption{Box plots representing the questionnaire results for the RPE (on the left) and perceived muscular fatigue for the deltoid group (on the right). For the RPE, 6 corresponds to the lowest score on the scale ("No exertion at all") and 11 corresponds to an RPE of "Light". The scale goes up to 20 ("Maximal exertion"). For perceived muscular fatigue, 1 corresponds to the lowest score on the Likert scale ("No Fatigue") and 7 corresponds to the highest score ("Extreme Fatigue").}
    \label{fig:rpe_fatigue}
\end{figure}

The RPE did not change significantly ($p = 0.125$) as well as the muscular fatigue for all muscle groups (deltoids $p = 0.859$; pectoralis major $p = 0.801$; trapezius $p = 0.397$; biceps $p = 0.691$; latissimus dorsi $p = 0.488$; triceps $p = 0.475$) across conditions (Fig. \ref{fig:rpe_fatigue}). 

\section{Discussion}
\subsection{Model Identification}
The developed model predicts a linear relationship between the desired output tension and the torque the TDU must apply to achieve that tension. During raising, the TDU motor must supply a higher torque to overcome the frictional losses in the transmission. On the contrary, when lowering, it is the arm of the user that must overcome the frictional losses in the transmission, hence the TDU motor must provide less torque accordingly. The two regression lines identified and used in the control strategy had high goodness of fit, validating the proposed model.

\subsection{System evaluation experiments on healthy participants}
\subsubsection{Assistive tension and torque tracking}

At the 50\% support condition, the average RMSE was 6.14N (17.4\%), 6.46N (18.2\%), and 8.17N (23.4\%), for velocities V1, V2, and V3, respectively. Georgarakis et al. reported a mean RMSE for tension tracking of 3.4N (16.1\%) (n = 5) when supporting 70\% of the weight of the upper arm using a load-cell to measure the cable tension and a tuned indirect force controller \citep{Georgarakis2020}. In terms of the RMSE errors, the controller developed by Georgarakis slightly outperforms the one developed in this paper. However, these results were achieved without a load-cell to explicitly regulate the error in cable tension.

Moreover, further optimizations are proposed that could improve the controller performance. The spike in cable tension error around 60\% of the movement cycle was a result of the transmission and motor stiction when the arm and cable changed direction. More complex modelling could capture the dynamics of stiction, or further mechanical optimizations could explore Bowden sheath-free cable transmissions (such as re-direct pulleys). A more thorough exploration of the sigmoid coupling parameters in Section \ref{sec:control} may also reduce the effect of this disturbance. Finally, the perceptions of the users should be evaluated to better characterize the requirements in terms of tension tracking accuracy.

So far no works have reported the magnitudes of errors for the supportive torques at the shoulder joint for upper-limb exosuits. These values may be used by the community as a reference and additionally may be compared between devices regardless of actuation modality. 

\subsubsection{Trajectory tracking and smoothness}

The RMSE in the trajectory tracking was not significantly affected by the presence of the suit and the increasing support. On the contrary, movement smoothness decreased significantly with increasing exosuit support, indicating that the exosuit slightly interfered with the natural motion of the user. As all participants were healthy, and the task of lifting a 0.5kg weight was relatively easy, it is possible that the higher supports disturbed movement smoothness since the assistance was not necessary to perform the task. 

\subsubsection{EMG}
The anterior deltoid, trapezius, and pectoralis major demonstrated the strongest decreases (-30\%, -38\%, and - 38\%, respectively) in muscular activity with 50\% gravity assistance of the exosuit. These results are comparable with those obtained by Georgarakis and colleagues \citep{Georgarakis2022} in which the activity of the anterior deltoid and trapezius was reduced by -42\% and -49\%, respectively, when supporting the limb with 70\% gravity assistance. Also Missiroli and coworkers \citep{Missiroli2022} achieved similar results with their hybrid occupational exoskeleton with a decrease of the activity of the anterior deltoid by -31\% when abducting the upper arm with passive support. As the exercise was a forward humeral elevation, one may expect to see decreased activity in these muscle groups as they are primarily responsible for supporting the arm against gravity. The activation of the posterior deltoid increased particularly (+32\%) during the lowering phase of the movement. This indicates that the suit was impeding the lowering motion and that the posterior deltoid activity had to increase to compensate for this. It is likely that the spikes in supportive torque (Fig. \ref{fig:single_part_results}) observed at the onset of lowering could be the cause of this increase in muscular activity. 

\subsubsection{Questionnaires}
Mild discomfort was perceived in the upper neck for all exosuit support levels, and mild to moderate discomfort in the right shoulder was perceived as the support level was increased. This discomfort is likely a result of the pressure of the shoulder cuff on the upper neck and shoulder during support. These results might have been exacerbated by the presence of the EMG system, as electrodes were placed on the trapezius and on the deltoid groups underneath the exosuit cuff. Nevertheless, more efforts should be invested in designing a comfortable shoulder cuff and exploring if the pressure could be tolerated by stakeholders for long-term use.

No significant changes in RPE and perceived muscular fatigue were found with increasing support from the exosuit. The results from both questionnaires indicate a potential floor-effect. A more physically demanding task might be necessary to elucidate differences in perceived exertion and muscular fatigue.

\subsubsection{Limitations and future improvements}
One limitation of this study lies in the fact that some of the electrodes for recording sEMG needed to be placed underneath the exosuit meaning that different pressures could have been applied to the electrodes for each condition. However, a higher contact pressure generally does not influence wet electrode impedance \citep{Taji2018}. Increased discomfort may occur resulting from the higher pressures.

Regarding the movement task, since the focus of this paper was on control and sensing, only a single movement was explored (forward humeral elevation). Even if this movement is one of the most relevant for ADLs \citep{Georgarakis2019}, future studies should include movements performed in the context of specific activities. Moreover, humeral rotation and wrist flexion were not enforced strictly, so the participants could choose how to orient the weight during the motions.

Finally, the textile part of the exosuit was one-size-fits-all. Even though straps were adjustable, this led to fit issues, particularly for participants smaller in stature. Future iterations of the exosuit should include different textile and cuff sizes to adapt to each user.

\section{Conclusion}
Soft exosuits have the potential to mature into a technology that can support a broad range of users with muscular weakness, both during in-clinic rehabilitation and for daily assistance at home. The proposed design and control strategy entailed mechanical improvements to the TDU and cable transmission along with a model-based tension controller tuned through a simple data-driven friction identification procedure. The exosuit successfully reduced activation in the muscles responsible for arm elevation, without affecting the humeral elevation tracking error in healthy subjects. This established the feasibility of sensorless cable tension control. This work may be leveraged to further the development of simpler cable-driven exosuits, bringing them closer to people with muscular weakness and difficulties performing activities of daily living.

\begin{Backmatter}

\paragraph{Acknowledgments}
The authors would like to thank Michi-Herold Nadig and Paul Schürmman for their support with integrating the electronics of the exosuit and Emanuele Bianchi for his support in developing the mannequin test bench. Additionally, the authors would like to express their thanks to the participants who volunteered their time and made this study possible.

\paragraph{Funding Statement}
This work was supported as a part of NCCR Robotics, a National Centre of Competence in Research, funded by the Swiss National Science Foundation (grant number 51NF40\_185543).

A travel grant was awarded to EB by IDEA League to support travel expenses.  

\paragraph{Competing Interests}
The authors declare that the research was conducted in the absence of any commercial or financial relationships that could be construed as a potential conflict of interest.

\paragraph{Data Availability Statement}
All raw data and scripts for processing and analysis are available upon request. Please contact: \url{adrian.esser@hest.ethz.ch}

\paragraph{Ethical Standards}
The protocol in this study was approved by the ethical committee of Politecnico di Milano (opinion n. 46/2022, 16/11/2022). Each participant read and signed a written informed consent prior to participation in the study.

\paragraph{Author Contributions}
%Please provide an author contributions statement using the CRediT taxonomy roles as a guide {\verb+\url{https://www.casrai.org/credit.html}+}. Conceptualization: A.A; A.B. Methodology: A.A; A.B. Data curation: A.C. Data visualisation: A.C. Writing original draft: A.A; A.B. All authors approved the final submitted draft.

\begin{table}[]
\begin{tabular}{lccccccc}
\textbf{CRediT Role}                  & \textbf{E.B.} & \textbf{A.E.} & \textbf{P.W.} & \textbf{M.G.} & \textbf{E.A.} & \textbf{A.P.} & \textbf{R.R.} \\
Conceptualization            &X&X& & & & & \\
Data Curation                &X&X& & & & & \\
Formal Analysis              &X&X& & & & & \\
Funding Acquisition          & & &X&X&X&X&X\\
Investigation                &X&X& & & & & \\
Methodology                  &X&X& &X&X& & \\
Project Administration       &X&X& & & & & \\
Resources                    &X&X& & & & & \\
Software                     &X&X& & & & & \\
Supervision                  & & &X&X&X&X&X\\
Validation                   &X&X& & & & & \\
Visualization                &X&X& & & & & \\
Writing (Original Draft)     &X&X& & & & & \\
Writing (Review and Editing) &X&X&X&X&X&X&X
\end{tabular}
\end{table}
All authors approved the final submitted draft.

\clearpage
\newpage

% For JDM please remove this \begin{thebibliography}...\end{thebibliography} list.
% Use biblatex-apa (see instructions in preamble) instead, and write \printbibliography here to print the reference list in APA7 style.
\printbibliography

\end{Backmatter}

\end{document}

\maketitle 

\section{Transmission model}
\label{supp:tranmission_model}
\begin{figure}[h]
    \centering
    \includegraphics[width=0.5\textwidth] {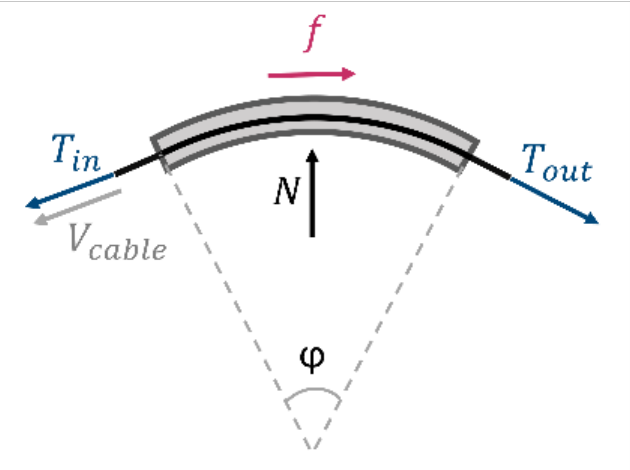}
    \caption{Free-body diagram of the exosuit cable inside the Bowden sheath. $T_{in}$ is the input tension on the TDU side, $T_{out}$ is the output tension on the limb side, $N$ is the normal reaction force acting on the cable, $f$ is the friction force, and $\phi$ is the total wrap angle of the sheath.}
    \label{fig:free-body}
\end{figure}

The free-body diagram of the Bowden-sheath cable system is shown in Figure \ref{fig:free-body}. The normal reaction force $N$ acting on the cable from the interaction with the sheath can be approximated as:
\begin{equation}
    N = (T_{in} + T_{out}) sin\left(\frac{\phi}{2}\right)
    \label{eq:normal_force}
\end{equation}
where $T_{in}$ is the input tendon tension on the motor side, $T_{out}$ is the output tendon tension acting on the limb, $\phi$ is the total wrap angle of the sheath.
From the equilibrium of the forces acting on the cable, it is possible to express the cable output tension as:
\begin{equation}
    T_{out} = T_{in} - f
    \label{eq:T_out}
\end{equation}
where $f$ is the friction force acting on the cable.
Substituting Eq. \ref{eq:T_out} in Eq. \ref{eq:normal_force}:
\begin{equation}
    N = (2T_{in} - f) sin\left(\frac{\phi}{2}\right)
    \label{eq:normal_force2}
\end{equation}
From the Coulomb friction model, the friction force, $f$, can be expressed as:
\begin{equation}
    f = sgn(V_{cable)} \mu N
    \label{eq:coulomb_friction}
\end{equation}
where $\mu$ is the Coulomb friction coefficient and $V_{cable}$ is the velocity of the cable relative to the sheath and considered positive if raising the arm (cable travels towards input), and negative if lowering the arm (cable travels towards output).
Substituting Eq. \ref{eq:normal_force2} in Eq. \ref{eq:coulomb_friction} it is obtained:
\begin{equation}
    f = sgn(V_{cable})\mu (2T_{in} - f) sin\left(\frac{\phi}{2}\right)
\end{equation}
and solving for $f$:
\begin{equation}
    f = sgn(V_{cable}) \frac{2 T_{in} \mu  sin\left(\frac{\phi}{2}\right)}{1 + sgn(V_{cable}) \mu sin\left(\frac{\phi}{2}\right)}
    \label{eq:friction_final}
\end{equation}

Considering inertial effects negligible, the output tension of a Bowden-sheath system can be computed as:
\begin{equation}
    \label{eq_supp:model_final}
    T_{out} = T_{in} \left(1 - sgn(V_{cable}) \frac{2 \mu  sin\left(\frac{\phi}{2}\right)}{1 + sgn(V_{cable}) \mu sin\left(\frac{\phi}{2}\right)}\right)
\end{equation}

\clearpage
\newpage
\section{Questionnaries}
\label{supp:questionnaires}
To evaluate discomfort participants filled out a 16-site modified Nordic Questionnaire with sites ranging from the upper neck to lower back and hands. Each site presented a 7-point Likert scale, ranging from "No Discomfort" to "Extreme Discomfort". To evaluate perceived exertion, the Borg Rate of Perceived Exertion (RPE) scale was used. Perceived muscular fatigue was evaluated using a 6-site (pectoralis major, deltoids, trapezius, biceps, triceps, latissimus dorsi) diagram with a 7-point Likert scale ranging from "No Fatigue" to "Extreme Fatigue".

\begin{center}
    
\includegraphics[width=0.7\textwidth]{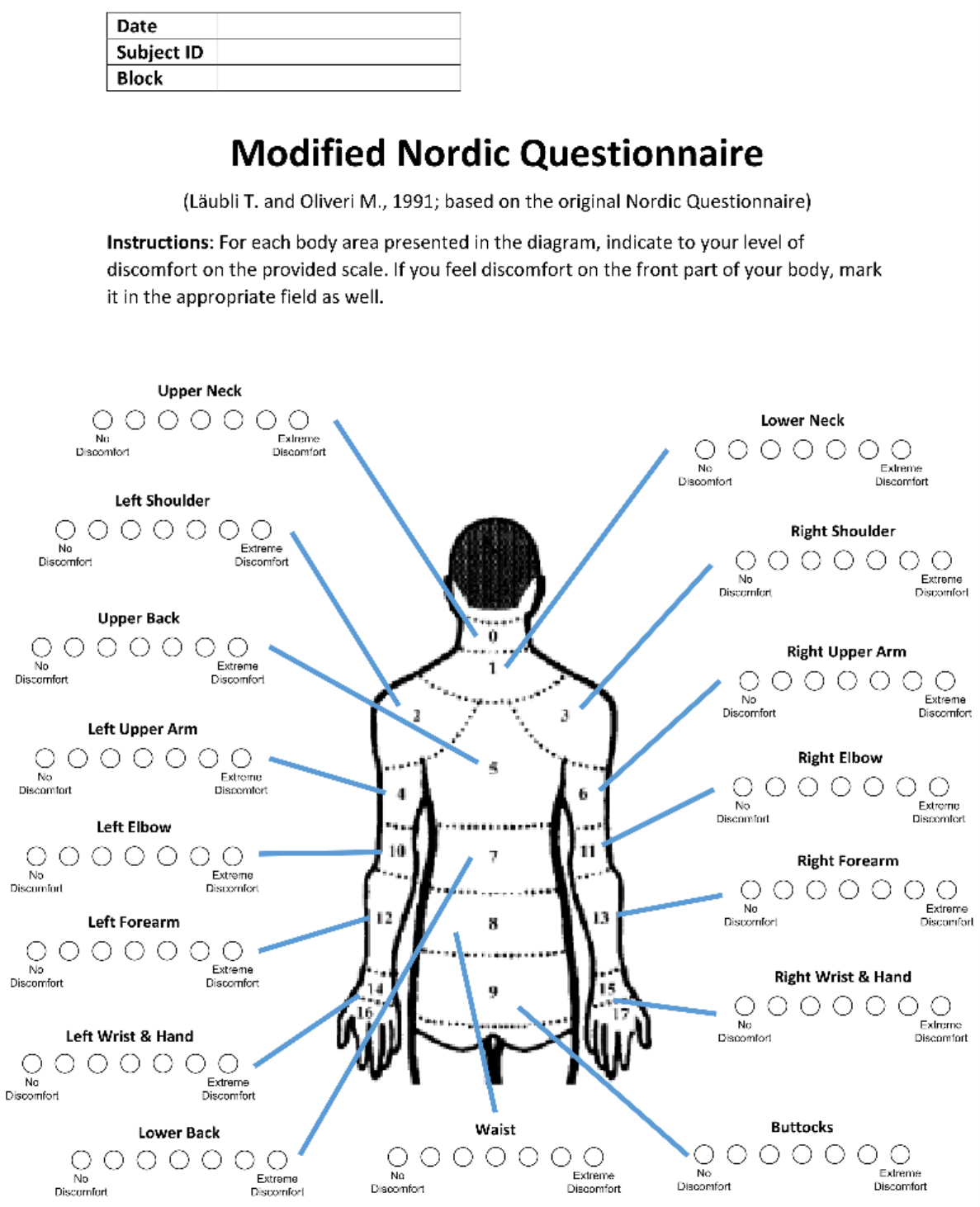}

\includegraphics[width=0.7\textwidth]{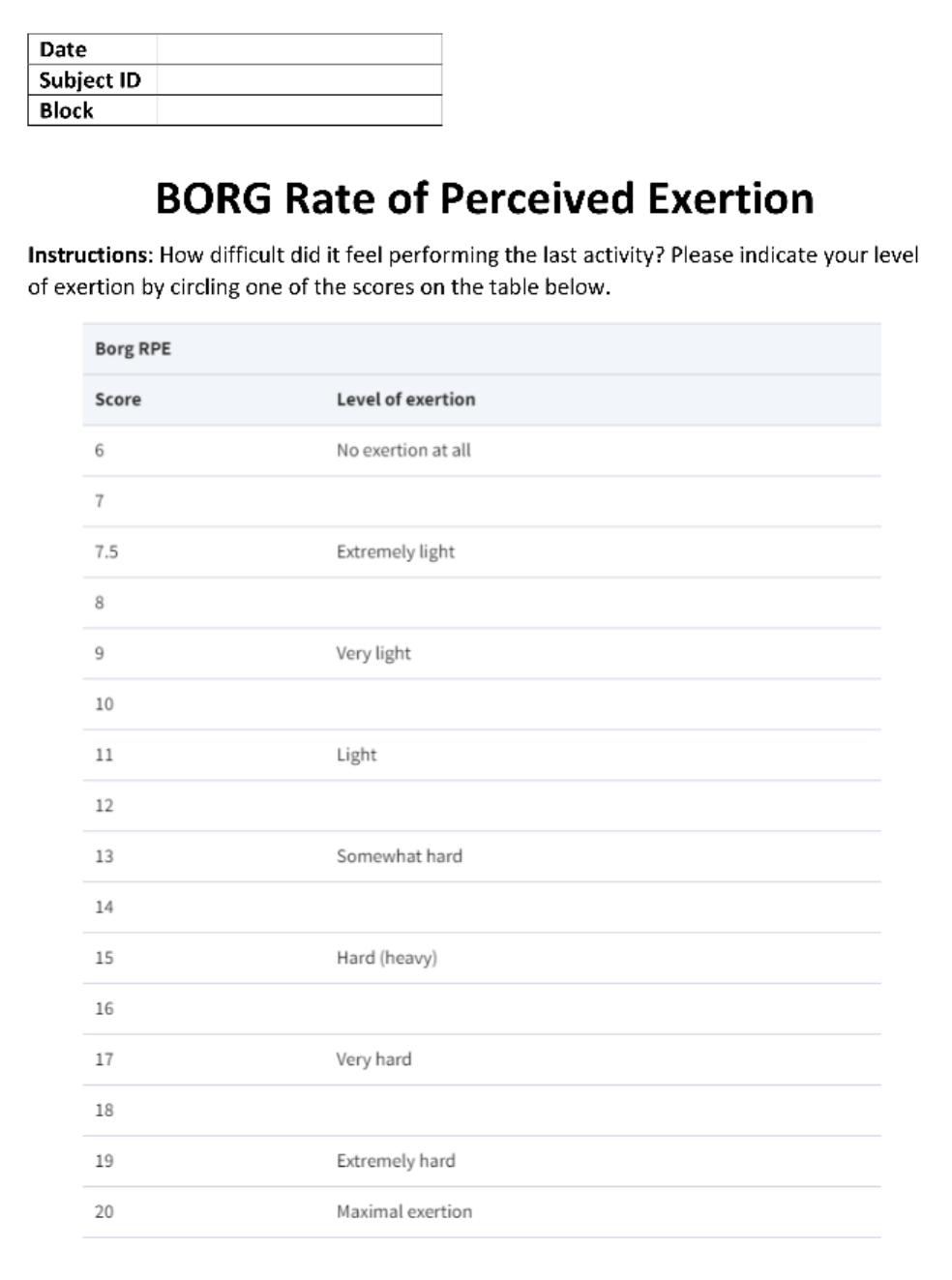}

\includegraphics[width=0.7\textwidth]{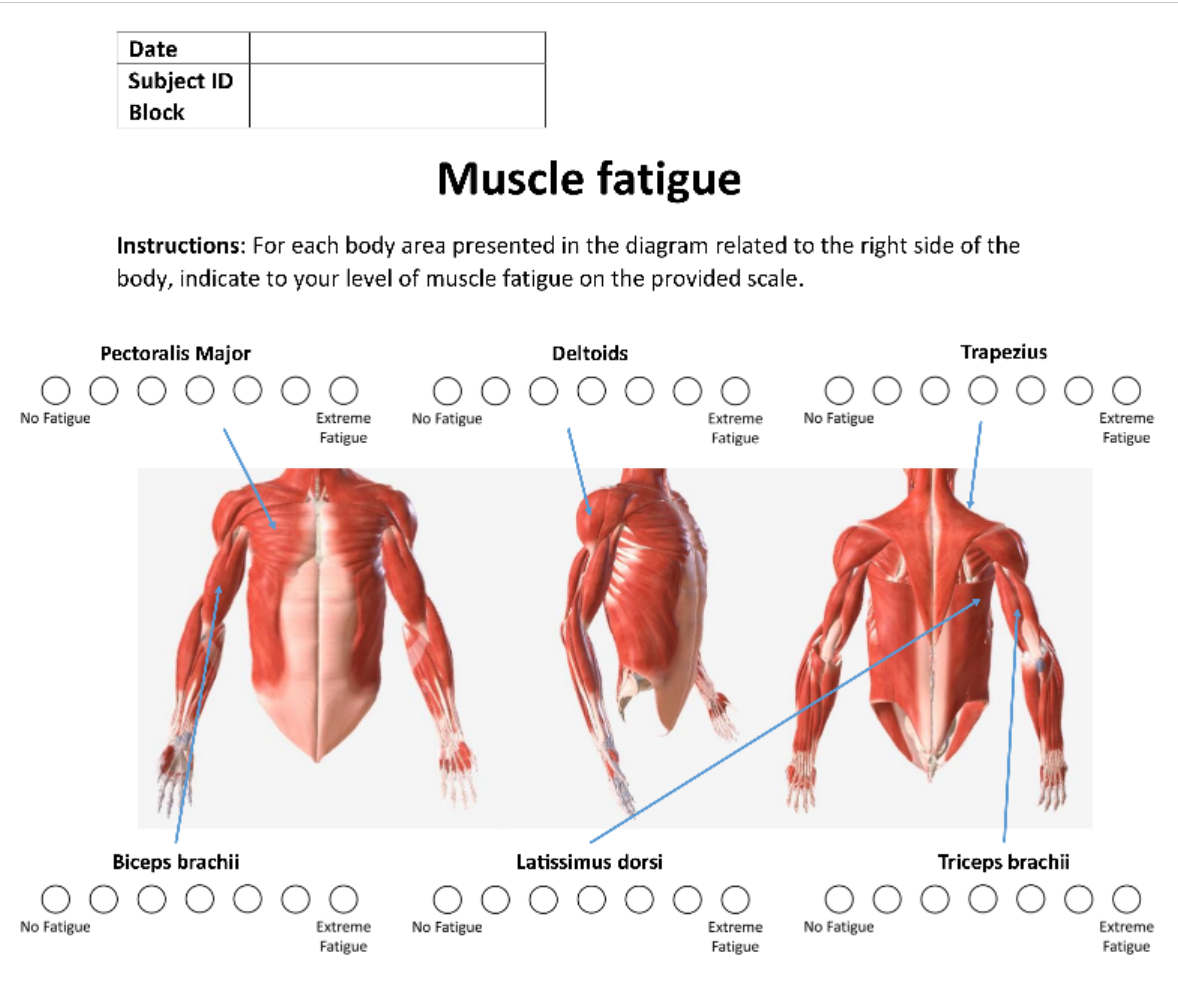}

\end{center}

\clearpage
\newpage
\section{Supplementary Data}

\subsection{Torque/tension tracking}

\begin{table}[h]
\centering
\resizebox{0.9\textwidth}{!}{%
\begin{tabular}{@{}cccccccccc@{}}
\toprule
\multicolumn{1}{l}{} & \multicolumn{3}{c}{\makecell{\textbf{V1} \\ ($v_{peak}$ = 60$\frac{^\circ}{s}$)}}           & \multicolumn{3}{c}{\makecell{\textbf{V2} \\ ($v_{peak}$ = 120$\frac{^\circ}{s}$)}}           & \multicolumn{3}{c}{\makecell{\textbf{V3} \\ ($v_{peak}$ = 180$\frac{^\circ}{s}$)}}           \\ \midrule
\makecell{\textbf{Pct. Err.} \\ \textbf{RMSE} \textbf{(\%)} } & \textbf{Entire} & \textbf{Raise} & \textbf{Lower} & \textbf{Entire} & \textbf{Raise} & \textbf{Lower} & \textbf{Entire} & \textbf{Raise} & \textbf{Lower} \\ 
\textbf{Pre} & 27.5 & 23.7 & 29.7 & 35.8 & 29.0 & 39.2 & 61.5 & 45.1 & 72.0 \\
\textbf{25\%} & 24.2 & 22.8 & 24.5 & 30.5 & 28.2 & 31.4 & 38.3 & 33.6 & 41.6 \\ 
\textbf{50\%} & 17.4 & 17.3 & 17.0 & 18.2 & 17.7 & 18.1 & 23.4 & 20.6 & 25.1 \\ \bottomrule
\end{tabular}%
}
\caption{\label{tab:tracking_pct_RMSE_combined} RMSE of the percentage error presented for all combinations of velocities and support conditions, averaged across all participants and repetitions. These values are identical for both tension and torque, as the kinematic model relating the two quantities is linear. The results are further split showing the average RMSE of the entire arm cycle, the raising portion, and the lowering portion.}
\end{table}

% Peak torque percentage errors and times with standard deviations, raising
\begin{table}[h]
\centering
\resizebox{0.9\textwidth}{!}{%
\begin{tabular}{@{}ccccccc@{}}
\toprule
\multicolumn{1}{l}{} & \multicolumn{2}{c}{\makecell{\textbf{V1} \\ ($v_{peak}$ = 60$\frac{^\circ}{s}$)}}           & \multicolumn{2}{c}{\makecell{\textbf{V2} \\ ($v_{peak}$ = 120$\frac{^\circ}{s}$)}}           & \multicolumn{2}{c}{\makecell{\textbf{V3} \\ ($v_{peak}$ = 180$\frac{^\circ}{s}$)}}           \\ \midrule
\multicolumn{1}{l}{} & \makecell{\textbf{Torque Pct.} \\ \textbf{Error (\%)}} & \makecell{\textbf{Normalized} \\ \textbf{Time (\%)}} & \makecell{\textbf{Torque Pct.} \\ \textbf{Error (\%)}} & \makecell{\textbf{Normalized} \\ \textbf{Time (\%)}} & \makecell{\textbf{Torque Pct.} \\ \textbf{Error (\%)}} & \makecell{\textbf{Normalized} \\ \textbf{Time (\%)}} \\
\textbf{Pre} & -29.0 (51.5) & 20.4 (18.6) & -32.3 (66.3) & 10.4 (10.8)  & -3.20 (112) & 17.4 (16.4) \\
\textbf{25\%} & -32.3 (44.3) & 17.2 (16.6) & -32.1 (64.5) & 10.6 (10.7) & -37.5 (70.1) & 12.0 (11.3) \\
\textbf{50\%} & -30.0 (24.8) & 18.2 (17.3) & -39.4 (16.9) & 7.80 (8.63) & -43.7 (23.9) & 8.27 (8.69) \\ \bottomrule
\end{tabular}%
}
\caption{\label{tab:peak_torque_percentage_errors_raising} Peak percentage errors for the assistive torques during the raising phase, averaged across all participants and repetitions with standard deviation in brackets. Normalized time and standard deviation for the peak values are reported as well.}
\end{table}

% Peak torque percentage errors and times with standard deviations, lowering
\begin{table}[h]
\centering
\resizebox{0.9\textwidth}{!}{%
\begin{tabular}{@{}ccccccc@{}}
\toprule
\multicolumn{1}{l}{} & \multicolumn{2}{c}{\makecell{\textbf{V1} \\ ($v_{peak}$ = 60$\frac{^\circ}{s}$)}}           & \multicolumn{2}{c}{\makecell{\textbf{V2} \\ ($v_{peak}$ = 120$\frac{^\circ}{s}$)}}           & \multicolumn{2}{c}{\makecell{\textbf{V3} \\ ($v_{peak}$ = 180$\frac{^\circ}{s}$)}}           \\ \midrule
\multicolumn{1}{l}{} & \makecell{\textbf{Torque Pct.} \\ \textbf{Error (\%)}} & \makecell{\textbf{Normalized} \\ \textbf{Time (\%)}} & \makecell{\textbf{Torque Pct.} \\ \textbf{Error (\%)}} & \makecell{\textbf{Normalized} \\ \textbf{Time (\%)}} & \makecell{\textbf{Torque Pct.} \\ \textbf{Error (\%)}} & \makecell{\textbf{Normalized} \\ \textbf{Time (\%)}} \\
\textbf{Pre} & 76.8 (54.1) & 68.4 (14.2) & 94.4 (71.0) & 60.5 (7.76) & 155 (90.3) & 61.2 (5.48) \\
\textbf{25\%} & 58.0 (53.0) & 74.0 (17.5) & 80.0 (63.0) & 61.4 (8.55) & 99.2 (56.5) & 62.3 (6.06) \\
\textbf{50\%} & 34.7 (29.9) & 74.1 (17.3) & 48.8 (25.3) & 63.6 (8.95) & 62.6 (29.2) & 63.6 (6.75) \\ \bottomrule
\end{tabular}%
}
\caption{\label{tab:peak_torque_percentage_errors_lowering} Peak percentage errors for the assistive torque from the exosuit during lowering phase, averaged across all participants and repetitions with standard deviation in brackets. Normalized time and standard deviation for the peak values are reported as well.}
\end{table}

\clearpage
\newpage
\subsection{Kinematics}
\begin{table}[h]
\centering
\resizebox{\textwidth}{!}{%
\begin{tabular}{clccccccc}
\hline
 & \multicolumn{1}{c}{} &  &  &  &  & \multirow{2}{*}{\begin{tabular}[c]{@{}c@{}}p-value\\ (C)\end{tabular}} & \multirow{2}{*}{\begin{tabular}[c]{@{}c@{}}p-value\\ (V)\end{tabular}} & \multirow{2}{*}{\begin{tabular}[c]{@{}c@{}}p-value\\ (C x V)\end{tabular}} \\
outcome & \multicolumn{1}{c}{speed} & no & pre & 25\% & 50\% &  &  &  \\ \hline
\multirow{3}{*}{RMSE (${}^\circ$)} & \rule{0pt}{1.0\normalbaselineskip}V1 & 4.755 (0.299) & 4.389 (0.301) & 4.514 (0.29) & 4.874 (0.309) & \multirow{3}{*}{0.335} & \multirow{3}{*}{\textbf{$<$ 0.001}} & \multirow{3}{*}{0.636} \\
 & V2 & 8.356 (0.584) & 7.769 (0.396) & 8.261 (0.462) & 8.345 (0.446) &  &  &  \\
 & V3 & 12.524 (0.868) & 13.586 (0.815) & 13.586 (0.862) & 13.888 (0.948) &  &  &  \\ \midrule
\multirow{3}{*}{SPARC (-)} & V1 & 2.76 (0.05) & 2.752 (0.041) & 2.789 (0.048) & 2.832 (0.055) & \multirow{3}{*}{\textbf{0.023}} & \multirow{3}{*}{\textbf{$<$ 0.001}} & \multirow{3}{*}{0.997} \\
 & V2 & 2.473 (0.046) & 2.476 (0.044) & 2.508 (0.047) & 2.554 (0.053) &  &  &  \\
 & V3 & 2.349 (0.042) & 2.368 (0.041) & 2.377 (0.041) & 2.416 (0.046) &  &  & \\ \hline
\end{tabular}%
}
\caption{\label{tab:shoulder_kinematics_statistical_results} Average shoulder position RMSE and SPARC, with standard deviation in brackets, across each velocity and condition are reported along with the p-values resulting from the GLMM considering condition, speed, and the interaction of condition and speed as fixed effects. The p-values for significant effects ($p<0.05$) are shown in bold. SPARC scores are traditionally always negative, but were converted to positive values for the statistical tests.}
\end{table}

\begin{table}[h]
\centering
\resizebox{\textwidth}{!}{%
\begin{tabular}{ccccccccccccc}
\hline
 & \multicolumn{2}{c}{pre - no} & \multicolumn{2}{c}{25\% - no} & \multicolumn{2}{c}{50\% - no} & \multicolumn{2}{c}{25\% - pre} & \multicolumn{2}{c}{50\% - pre} & \multicolumn{2}{c}{50\% - 25\%} \\
outcome & change & p-value & change & p-value & change & p-value & change & p-value & change & p-value & change & p-value \\ \hline
\rule{0pt}{1.0\normalbaselineskip}RMSE (${}^\circ$) & N/A & N/A & N/A & N/A & N/A & N/A & N/A & N/A & N/A & N/A & N/A & N/A \\
SPARC (-) & \begin{tabular}[c]{@{}c@{}}0.005\\ (0.021)\end{tabular} & 0.809 & \begin{tabular}[c]{@{}c@{}}0.031\\ (0.023)\end{tabular} & 0.552 & \begin{tabular}[c]{@{}c@{}}0.073\\ (0.027)\end{tabular} & \textbf{0.04} & \begin{tabular}[c]{@{}c@{}}0.025\\ (0.021)\end{tabular} & 0.552 & \begin{tabular}[c]{@{}c@{}}0.068\\ (0.025)\end{tabular} & \textbf{0.04} & \begin{tabular}[c]{@{}c@{}}0.042\\ (0.026)\end{tabular} & 0.42 \\ \hline
\end{tabular}%
}
\caption{\label{tab:shoulder_kinematics_condition_comparisons} Pairwise comparisons across conditions for SPARC. Post-hoc pairwise comparisons for shoulder position RMSE were not performed since the effect of condition was not significant. The p-values for significant comparisons ($p<0.05$) are shown in bold. A positive change in SPARC means that the first condition was less smooth than the second.}
\end{table}

\begin{table}[h]
\centering
\resizebox{\textwidth}{!}{%
\begin{tabular}{ccccccc}
\hline
 & \multicolumn{2}{c}{V1-V2} & \multicolumn{2}{c}{V1-V3} & \multicolumn{2}{c}{V2-V3} \\
outcome & change & p-value & change & p-value & change & p-value \\ \hline 
\rule{0pt}{1.0\normalbaselineskip}RMSE (${}^\circ$) & -3.55   (0.249) & \textbf{$<$ 0.001} & -8.756 (0.511) & \textbf{$<$ 0.001} & -5.207 (0.431) & \textbf{$<$ 0.001} \\
SPARC (-) & 0.28 (0.022) & \textbf{$<$ 0.001} & 0.406 (0.02) & \textbf{$<$ 0.001} & 0.125 (0.021) & \textbf{$<$ 0.001} \\ \hline
\end{tabular}%
}
\caption{\label{tab:shoulder_kinematics_velocity_comparisons} Pairwise comparisons across velocities for changes in RMSE and SPARC. The p-values for significant comparisons ($p<0.05$) are shown in bold. A negative change in RMSE means that the first velocity had greater tracking errors than the second. A positive change in SPARC means that the first condition was less smooth than the second.}
\end{table}

\clearpage
\newpage
\subsection{EMG}
\begin{table}[h]
\centering
\resizebox{\textwidth}{!}{%
\begin{tabular}{@{}ccccccccc@{}}
\toprule
           &    &                 &               &               &               & p-value          & p-value   & p-value \\
Outcome    &    & no              & pre           & 25\%          & 50\%          & (C)              & (V)       & (C x V) \\ \midrule
           & V1 & 0.435   (0.037) & 0.358 (0.028) & 0.324 (0.025) & 0.285 (0.027) & \textbf{}        & \textbf{} &         \\
Ant. Delt. & V2 & 0.459 (0.039)   & 0.380 (0.026) & 0.353 (0.027) & 0.307 (0.024) & \textbf{$<$ 0.001} & 0.116     & 0.996   \\
           & V3 & 0.437 (0.037)   & 0.375 (0.027) & 0.338 (0.027) & 0.305 (0.027) &                  &           &         \\ \midrule
           & V1 & 0.392 (0.043)   & 0.401 (0.037) & 0.384 (0.036) & 0.327 (0.031) & \textbf{}        & \textbf{} &         \\
Med. Delt.  & V2 & 0.419 (0.045) & 0.445 (0.042) & 0.440 (0.042) & 0.369 (0.036) & \textbf{$<$ 0.001} & \textbf{$<$ 0.001} & 0.846 \\
           & V3 & 0.434 (0.049)   & 0.493 (0.047) & 0.457 (0.046) & 0.427 (0.043) &                  &           &         \\ \midrule
           & V1 & 0.415 (0.035)   & 0.447 (0.036) & 0.439 (0.037) & 0.480 (0.046) &                  & \textbf{} &         \\
Post. Delt. & V2 & 0.454 (0.038) & 0.482 (0.035) & 0.492 (0.035) & 0.507 (0.037) & \textbf{0.005}   & \textbf{0.002}   & 0.837 \\
           & V3 & 0.461 (0.039)   & 0.499 (0.035) & 0.499 (0.039) & 0.576 (0.046) &                  &           &         \\ \midrule
           & V1 & 0.485 (0.040)   & 0.619 (0.063) & 0.573 (0.058) & 0.519 (0.051) & \textbf{}        & \textbf{} &         \\
Biceps     & V2 & 0.438 (0.034)   & 0.478 (0.035) & 0.492 (0.040) & 0.446 (0.038) & \textbf{0.006}   & 0.754     & 0.754   \\
           & V3 & 0.353 (0.029)   & 0.366 (0.028) & 0.389 (0.031) & 0.358 (0.039) &                  &           &         \\ \midrule
           & V1 & 0.471 (0.055)   & 0.445 (0.043) & 0.434 (0.039) & 0.426 (0.046) & \textbf{}        & \textbf{} &         \\
Triceps    & V2 & 0.492 (0.053)   & 0.487 (0.040) & 0.475 (0.041) & 0.425 (0.036) & 0.061            & 0.87      & 0.87    \\
           & V3 & 0.492 (0.057)   & 0.483 (0.043) & 0.454 (0.041) & 0.445 (0.039) &                  &           &         \\ \midrule
           & V1 & 0.491 (0.046)   & 0.400 (0.032) & 0.380 (0.032) & 0.311 (0.029) & \textbf{}        & \textbf{} &         \\
Trapezius  & V2 & 0.474 (0.042)   & 0.391 (0.029) & 0.354 (0.027) & 0.289 (0.024) & \textbf{$<$ 0.001} & 0.985     & 0.985   \\
           & V3 & 0.422 (0.039)   & 0.325 (0.028) & 0.307 (0.023) & 0.254 (0.022) &                  &           &         \\ \midrule
           & V1 & 0.395 (0.024)   & 0.387 (0.025) & 0.386 (0.028) & 0.360 (0.034) & \textbf{}        & \textbf{} &         \\
Lat. Dorsi & V2 & 0.424 (0.023)   & 0.428 (0.027) & 0.437 (0.032) & 0.402 (0.033) & 0.838            & 0.838     & 0.838   \\
           & V3 & 0.460 (0.028)   & 0.473 (0.034) & 0.490 (0.036) & 0.495 (0.043) &                  &           &         \\ \midrule
           & V1 & 0.548 (0.075)   & 0.431 (0.051) & 0.401 (0.045) & 0.347 (0.041) & \textbf{}        & \textbf{} &         \\
Pectoralis & V2 & 0.511 (0.067)   & 0.398 (0.041) & 0.376 (0.041) & 0.334 (0.039) & \textbf{$<$ 0.001} & 0.999     & 0.999   \\
           & V3 & 0.442 (0.056)   & 0.360 (0.041) & 0.335 (0.043) & 0.291 (0.035) &                  &           &         \\ \bottomrule
\end{tabular}%
}
\caption{Average normalized iEMG for the entire arm cycle, with standard deviation in brackets, across each velocity (V) and condition (C). The p-values are given considering condition, speed, and the interaction of condition and speed as fixed effects. The p-values for significant effects ($p<0.05$) are shown in bold.}
\label{tab:iEMG_whole_mean}
\end{table}

% \usepackage{booktabs}
% \usepackage{graphicx}
\begin{table}[h]
\centering
\resizebox{\textwidth}{!}{%
\begin{tabular}{@{}ccccccccc@{}}
\toprule
            &    &               &               &               &               & p-value         & p-value         & p-value \\
Outcome     &    & no            & pre           & 25\%          & 50\%          & (C)             & (V)             & (C x V) \\ \midrule
            & V1 & 0.548 (0.045) & 0.459 (0.035) & 0.425 (0.033) & 0.369 (0.038) & \textbf{}       & \textbf{}       &         \\
Ant. Delt.  & V2 & 0.629 (0.053) & 0.532 (0.035) & 0.502 (0.038) & 0.449 (0.035) & \textbf{$<$ 0.001} & \textbf{$<$ 0.001} & 0.989   \\
            & V3 & 0.622 (0.053) & 0.530 (0.037) & 0.479 (0.037) & 0.452 (0.040) &                 &                 &         \\ \midrule
            & V1 & 0.486 (0.054) & 0.494 (0.047) & 0.477 (0.046) & 0.401 (0.042) & \textbf{}       & \textbf{}       &         \\
Med. Delt.  & V2 & 0.583 (0.064) & 0.611 (0.058) & 0.614 (0.058) & 0.524 (0.050) & \textbf{$<$ 0.001} & \textbf{$<$ 0.001} & 0.859   \\
            & V3 & 0.639 (0.076) & 0.705 (0.067) & 0.657 (0.066) & 0.620 (0.062) &                 &                 &         \\ \midrule
            & V1 & 0.488 (0.045) & 0.503 (0.043) & 0.491 (0.043) & 0.483 (0.041) &                 & \textbf{}       &         \\
Post. Delt. & V2 & 0.580 (0.054) & 0.611 (0.049) & 0.616 (0.049) & 0.602 (0.049) & 0.695           & \textbf{$<$ 0.001} & 0.928   \\
            & V3 & 0.612 (0.059) & 0.652 (0.050) & 0.642 (0.054) & 0.686 (0.061) &                 &                 &         \\ \midrule
            & V1 & 0.537 (0.029) & 0.686 (0.067) & 0.647 (0.061) & 0.569 (0.051) & \textbf{}       & \textbf{}       &         \\
Biceps      & V2 & 0.517 (0.024) & 0.560 (0.032) & 0.586 (0.042) & 0.542 (0.048) & \textbf{0.009}  & \textbf{$<$ 0.001} & 0.55    \\
            & V3 & 0.432 (0.024) & 0.427 (0.030) & 0.465 (0.035) & 0.433 (0.052) &                 &                 &         \\ \midrule
            & V1 & 0.543 (0.069) & 0.509 (0.056) & 0.497 (0.051) & 0.466 (0.056) & \textbf{}       & \textbf{}       &         \\
Triceps     & V2 & 0.610 (0.071) & 0.616 (0.057) & 0.605 (0.058) & 0.531 (0.051) & \textbf{0.01}   & \textbf{$<$ 0.001} & 0.974   \\
            & V3 & 0.644 (0.079) & 0.632 (0.060) & 0.590 (0.058) & 0.550 (0.054) &                 &                 &         \\ \midrule
            & V1 & 0.595 (0.049) & 0.509 (0.038) & 0.483 (0.038) & 0.367 (0.030) & \textbf{}       & \textbf{}       &         \\
Trapezius   & V2 & 0.620 (0.050) & 0.551 (0.037) & 0.492 (0.033) & 0.395 (0.030) & \textbf{$<$ 0.001} & \textbf{0.003}  & 0.953   \\
            & V3 & 0.574 (0.047) & 0.465 (0.036) & 0.446 (0.029) & 0.356 (0.028) &                 &                 &         \\ \midrule
            & V1 & 0.463 (0.026) & 0.447 (0.028) & 0.449 (0.032) & 0.377 (0.030) & \textbf{}       & \textbf{}       &         \\
Lat. Dorsi  & V2 & 0.537 (0.028) & 0.525 (0.030) & 0.538 (0.041) & 0.458 (0.037) & \textbf{0.003}  & \textbf{$<$ 0.001} & 0.993   \\
            & V3 & 0.593 (0.036) & 0.580 (0.038) & 0.600 (0.042) & 0.527 (0.042) &                 &                 &         \\ \midrule
            & V1 & 0.598 (0.065) & 0.485 (0.049) & 0.463 (0.044) & 0.377 (0.037) & \textbf{}       & \textbf{}       &         \\
Pectoralis  & V2 & 0.595 (0.060) & 0.474 (0.041) & 0.443 (0.039) & 0.379 (0.035) & \textbf{$<$ 0.001} & \textbf{$<$ 0.001} & 0.995   \\
            & V3 & 0.536 (0.054) & 0.413 (0.039) & 0.385 (0.043) & 0.324 (0.032) &                 &                 &         \\ \bottomrule
\end{tabular}%
}
\caption{Average normalized iEMG for the raising phase, with standard deviation in brackets, across each velocity (V) and condition (C). The p-values are given considering condition, speed, and the interaction of condition and speed as fixed effects. The p-values for significant effects ($p<0.05$) are shown in bold.}
\label{tab:iEMG_raise_mean}
\end{table}

\begin{table}[h]
\centering
\resizebox{\textwidth}{!}{%
\begin{tabular}{@{}ccccccccc@{}}
\toprule
            &    &                 &               &               &               & p-value          & p-value          & p-value \\
Outcome     &    & no              & pre           & 25\%          & 50\%          & (C)              & (V)              & (C x V) \\ \midrule
            & V1 & 0.320   (0.032) & 0.259 (0.024) & 0.226 (0.021) & 0.200 (0.024) & \textbf{}        & \textbf{}        &         \\
Ant. Delt.  & V2 & 0.294 (0.028)   & 0.241 (0.020) & 0.214 (0.019) & 0.174 (0.018) & \textbf{$<$ 0.001} & \textbf{$<$ 0.001} & 0.992   \\
            & V3 & 0.261 (0.026)   & 0.221 (0.019) & 0.196 (0.019) & 0.162 (0.018) &                  &                  &         \\ \midrule
            & V1 & 0.296 (0.034)   & 0.311 (0.030) & 0.293 (0.028) & 0.252 (0.025) & \textbf{}        & \textbf{}        &         \\
Med. Delt.  & V2 & 0.264 (0.028)   & 0.292 (0.028) & 0.279 (0.027) & 0.225 (0.024) & \textbf{$<$ 0.001} & \textbf{0.006}   & 0.896   \\
            & V3 & 0.242 (0.026)   & 0.283 (0.028) & 0.258 (0.027) & 0.237 (0.027) &                  &                  &         \\ \midrule
            & V1 & 0.341 (0.033)   & 0.388 (0.036) & 0.381 (0.035) & 0.455 (0.057) &                  & \textbf{}        &         \\
Post. Delt. & V2 & 0.330 (0.030)   & 0.357 (0.029) & 0.369 (0.030) & 0.403 (0.037) & \textbf{$<$ 0.001} & 0.171            & 0.788   \\
            & V3 & 0.318 (0.031)   & 0.344 (0.029) & 0.352 (0.033) & 0.447 (0.048) &                  &                  &         \\ \midrule
            & V1 & 0.432 (0.054)   & 0.555 (0.070) & 0.507 (0.064) & 0.472 (0.062) & \textbf{}        & \textbf{}        &         \\
Biceps      & V2 & 0.357 (0.042)   & 0.399 (0.041) & 0.402 (0.043) & 0.359 (0.038) & \textbf{0.006}   & \textbf{$<$ 0.001} & 0.934   \\
            & V3 & 0.275 (0.032)   & 0.302 (0.031) & 0.310 (0.033) & 0.279 (0.032) &                  &                  &         \\ \midrule
            & V1 & 0.399 (0.050)   & 0.380 (0.038) & 0.370 (0.035) & 0.380 (0.045) & \textbf{}        & \textbf{}        &         \\
Triceps     & V2 & 0.374 (0.044)   & 0.361 (0.032) & 0.348 (0.032) & 0.316 (0.030) & 0.325            & \textbf{0.006}   & 0.683   \\
            & V3 & 0.345 (0.047)   & 0.327 (0.032) & 0.312 (0.030) & 0.325 (0.033) &                  &                  &         \\ \midrule
            & V1 & 0.392 (0.055)   & 0.291 (0.034) & 0.275 (0.032) & 0.256 (0.041) & \textbf{}        & \textbf{}        &         \\
Trapezius   & V2 & 0.334 (0.049)   & 0.236 (0.026) & 0.218 (0.025) & 0.185 (0.024) & \textbf{$<$ 0.001} & \textbf{$<$ 0.001} & 0.896   \\
            & V3 & 0.273 (0.039)   & 0.179 (0.024) & 0.160 (0.019) & 0.148 (0.019) &                  &                  &         \\ \midrule
            & V1 & 0.327 (0.031)   & 0.325 (0.026) & 0.320 (0.028) & 0.336 (0.041) & \textbf{}        & \textbf{}        &         \\
Lat. Dorsi  & V2 & 0.314 (0.025)   & 0.337 (0.029) & 0.341 (0.030) & 0.343 (0.035) & 0.085            & \textbf{0.003}   & 0.515   \\
            & V3 & 0.334 (0.028)   & 0.368 (0.039) & 0.379 (0.037) & 0.452 (0.053) &                  &                  &         \\ \midrule
            & V1 & 0.504 (0.092)   & 0.373 (0.056) & 0.340 (0.050) & 0.308 (0.047) & \textbf{}        & \textbf{}        &         \\
Pectoralis  & V2 & 0.431 (0.076)   & 0.319 (0.045) & 0.302 (0.044) & 0.281 (0.044) & \textbf{$<$ 0.001} & \textbf{$<$ 0.001} & 0.973   \\
            & V3 & 0.355 (0.059)   & 0.297 (0.045) & 0.277 (0.044) & 0.248 (0.040) &                  &                  &         \\ \bottomrule
\end{tabular}%
}
\caption{Average normalized iEMG for the lowering phase, with standard deviation in brackets, across each velocity (V) and condition (C). The p-values are given considering condition, speed, and the interaction of condition and speed as fixed effects. The p-values for significant effects ($p<0.05$) are shown in bold.}
\label{tab:iEMG_lower_mean}
\end{table}

% \usepackage{graphicx}
\begin{landscape}
\begin{table}[]
\centering
\resizebox{\columnwidth}{!}{%
\begin{tabular}{|l|rrr|rrr|rrr|rrr|rrr|rrr|}
\hline
\multicolumn{1}{|c|}{Whole} &
  \multicolumn{3}{c|}{pre - no} &
  \multicolumn{3}{c|}{25\% - no} &
  \multicolumn{3}{c|}{50\% - no} &
  \multicolumn{3}{c|}{25\% - pre} &
  \multicolumn{3}{c|}{50\% - pre} &
  \multicolumn{3}{c|}{50\% - 25\%} \\
\multicolumn{1}{|c|}{Muscle} &
  \multicolumn{2}{c}{change} &
  \multicolumn{1}{c|}{p-value} &
  \multicolumn{2}{c}{change} &
  \multicolumn{1}{c|}{p-value} &
  \multicolumn{2}{c}{change} &
  \multicolumn{1}{c|}{p-value} &
  \multicolumn{2}{c}{change} &
  \multicolumn{1}{c|}{p-value} &
  \multicolumn{2}{c}{change} &
  \multicolumn{1}{c|}{p-value} &
  \multicolumn{2}{c}{change} &
  \multicolumn{1}{c|}{p-value} \\ \hline
Ant. delt. &
  -0.072 & (-16.3\%) &
  \textbf{$<$ 0.001} &
  -0.105 & (-23.8\%) &
  \textbf{$<$ 0.001} &
  -0.145 & (-32.6\%) &
  \textbf{$<$ 0.001} &
  -0.033 & (-9.0\%) &
  \textbf{0.003} &
  -0.073 & (-19.5\%) &
  \textbf{$<$ 0.001} &
  -0.039 & (-11.6\%) &
  \textbf{0.003} \\
Med. delt. &
  0.030 & (7.3\%) &
  0.359 &
  0.011 & (2.7\%) &
  0.566 &
  -0.042 & (-10.2\%) &
  0.131 &
  -0.019 & (-4.2\%) &
  0.398 &
  -0.072 & (-16.3\%) &
  \textbf{$<$ 0.001} &
  -0.054 & (-12.6\%) &
  \textbf{0.003} \\
Post. delt. &
  0.032 & (7.3\%) &
  0.184 &
  0.033 & (7.4\%) &
  0.184 &
  0.076 & (17.2\%) &
  \textbf{0.003} &
  0.001 & (0.2\%) &
  0.963 &
  0.044 & (9.2\%) &
  0.110 &
  0.043 & (9.1\%) &
  0.136 \\
Biceps &
  0.055 & (13.1\%) &
  \textbf{0.005} &
  0.057 & (13.6\%) &
  \textbf{0.006} &
  0.014 & (3.3\%) &
  0.529 &
  0.002 & (0.5\%) &
  0.924 &
  -0.041 & (-8.6\%) &
  0.093 &
  -0.043 & (-9.1\%) &
  0.087 \\
Triceps &
  -0.013 & (-2.8\%) &
  0.675 &
  -0.030 & (-6.3\%) &
  0.675 &
  -0.053 & (-10.9\%) &
  0.224 &
  -0.017 & (-3.6\%) &
  0.675 &
  -0.040 & (-8.4\%) &
  0.107 &
  -0.023 & (-5.0\%) &
  0.650 \\
Lat. dorsi &
  0.002 & (0.5\%) &
  1.000 &
  0.010 & (2.2\%) &
  1.000 &
  -0.010 & (-2.4\%) &
  1.000 &
  0.008 & (1.8\%) &
  1.000 &
  -0.012 & (-2.9\%) &
  1.000 &
  -0.020 & (-4.6\%) &
  1.000 \\
Pectoralis &
  -0.103 & (-20.6\%) &
  \textbf{0.001} &
  -0.128 & (-25.8\%) &
  \textbf{$<$ 0.001} &
  -0.175 & (-35.1\%) &
  \textbf{$<$ 0.001} &
  -0.026 & (-6.5\%) &
  0.104 &
  -0.072 & (-18.2\%) &
  \textbf{$<$ 0.001} &
  -0.046 & (-12.6\%) &
  \textbf{0.013} \\
Trapezius &
  -0.090 & (-19.6\%) &
  \textbf{$<$ 0.001} &
  -0.116 & (-25.1\%) &
  \textbf{$<$ 0.001} &
  -0.178 & (-38.5\%) &
  \textbf{$<$ 0.001} &
  -0.025 & (-6.8\%) &
  \textbf{0.041} &
  -0.087 & (-23.5\%) &
  \textbf{$<$ 0.001} &
  -0.062 & (-17.9\%) &
  \textbf{$<$ 0.001} \\ \bottomrule
\end{tabular}%
}
\caption{Pairwise comparisons across conditions for changes in iEMG for the entire movement cycle. The p-values for significant comparisons ($p<0.05$) are shown in bold. A negative change in iEMG means that the muscle activation was higher in the second condition.}
\label{tab:pairwise_whole}
\end{table}

% \usepackage{graphicx}

\begin{table}[]
\centering
\resizebox{\columnwidth}{!}{%
\begin{tabular}{|l|rrr|rrr|rrr|rrr|rrr|rrr|}
\hline
\multicolumn{1}{|c|}{Raising} &
  \multicolumn{3}{c|}{pre - no} &
  \multicolumn{3}{c|}{25\% - no} &
  \multicolumn{3}{c|}{50\% - no} &
  \multicolumn{3}{c|}{25\% - pre} &
  \multicolumn{3}{c|}{50\% - pre} &
  \multicolumn{3}{c|}{50\% - 25\%} \\
\multicolumn{1}{|c|}{Muscle} &
  \multicolumn{2}{c}{change} &
  \multicolumn{1}{c|}{p-value} &
  \multicolumn{2}{c}{change} &
  \multicolumn{1}{c|}{p-value} &
  \multicolumn{2}{c}{change} &
  \multicolumn{1}{c|}{p-value} &
  \multicolumn{2}{c}{change} &
  \multicolumn{1}{c|}{p-value} &
  \multicolumn{2}{c}{change} &
  \multicolumn{1}{c|}{p-value} &
  \multicolumn{2}{c}{change} &
  \multicolumn{1}{c|}{p-value} \\ \hline
Ant. delt. &
  -0.093 & (-15.5\%) &
  \textbf{$<$ 0.001} &
  -0.131 & (-21.9\%) &
  \textbf{$<$ 0.001} &
  -0.177 & (-29.6\%) &
  \textbf{$<$ 0.001} &
  -0.038 & (-7.6\%) &
  \textbf{0.019} &
  -0.084 & (-16.7\%) &
  \textbf{$<$ 0.001} &
  -0.046 & (-9.8\%) &
  \textbf{0.019} \\
Med. delt. &
  0.031 & (5.5\%) &
  0.776 &
  0.012 & (2.1\%) &
  0.776 &
  -0.059 & (-10.4\%) &
  0.168 &
  -0.020 & (-3.3\%) &
  0.776 &
  -0.090 & (-15.1\%) &
  \textbf{$<$ 0.001} &
  -0.071 & (-12.2\%) &
  \textbf{0.010} \\
Post. delt. &
  0.028 & (4.9\%) &
  1.000 &
  0.021 & (3.8\%) &
  1.000 &
  0.027 & (4.8\%) &
  1.000 &
  -0.006 & (-1.1\%) &
  1.000 &
  -0.001 & (-0.1\%) &
  1.000 &
  0.006 & (1.0\%) &
  1.000 \\
Biceps &
  0.055 & (11.1\%) &
  \textbf{0.019} &
  0.068 & (13.8\%) &
  \textbf{0.007} &
  0.018 & (3.7\%) &
  0.529 &
  0.013 & (2.4\%) &
  0.666 &
  -0.036 & (-6.7\%) &
  0.287 &
  -0.050 & (-8.9\%) &
  0.162 \\
Triceps &
  -0.014 & (-2.3\%) &
  0.806 &
  -0.035 & (-5.9\%) &
  0.806 &
  -0.083 & (-13.9\%) &
  0.074 &
  -0.021 & (-3.7\%) &
  0.806 &
  -0.069 & (-11.8\%) &
  \textbf{0.016} &
  -0.047 & (-8.5\%) &
  0.120 \\
Lat. dorsi &
  -0.014 & (-2.7\%) &
  1.000 &
  -0.003 & (-0.6\%) &
  1.000 &
  -0.078 & (-14.8\%) &
  \textbf{0.001} &
  0.011 & (2.2\%) &
  1.000 &
  -0.064 & (-12.5\%) &
  \textbf{0.014} &
  -0.075 & (-14.3\%) &
  \textbf{0.013} \\
Pectoralis &
  -0.119 & (-20.7\%) &
  \textbf{$<$ 0.001} &
  -0.146 & (-25.4\%) &
  \textbf{0.000} &
  -0.216 & (-37.6\%) &
  \textbf{$<$ 0.001} &
  -0.027 & (-5.9\%) &
  0.150 &
  -0.097 & (-21.3\%) &
  \textbf{$<$ 0.001} &
  -0.070 & (-16.3\%) &
  \textbf{$<$ 0.001} \\
Trapezius &
  -0.089 & (-14.9\%) &
  \textbf{0.001} &
  -0.123 & (-20.6\%) &
  \textbf{0.000} &
  -0.223 & (-37.5\%) &
  \textbf{$<$ 0.001} &
  -0.034 & (-6.7\%) &
  \textbf{0.044} &
  -0.135 & (-26.6\%) &
  \textbf{$<$ 0.001} &
  -0.101 & (-21.3\%) &
  \textbf{$<$ 0.001} \\ \bottomrule
\end{tabular}%
}
\caption{Pairwise comparisons across conditions for changes in iEMG for the raising phase. The p-values for significant comparisons ($p<0.05$) are shown in bold. A negative change in iEMG means that the muscle activation was higher in the second condition.}
\label{tab:pairwise_raising}
\end{table}

% \usepackage{graphicx}

\begin{table}[]
\centering
\resizebox{\columnwidth}{!}{%
\begin{tabular}{|l|rrr|rrr|rrr|rrr|rrr|rrr|}
\hline
\multicolumn{1}{|c|}{Lowering} &
  \multicolumn{3}{c|}{pre - no} &
  \multicolumn{3}{c|}{25\% - no} &
  \multicolumn{3}{c|}{50\% - no} &
  \multicolumn{3}{c|}{25\% - pre} &
  \multicolumn{3}{c|}{50\% - pre} &
  \multicolumn{3}{c|}{50\% - 25\%} \\
\multicolumn{1}{|c|}{Muscle} &
  \multicolumn{2}{c}{change} &
  \multicolumn{1}{c|}{p-value} &
  \multicolumn{2}{c}{change} &
  \multicolumn{1}{c|}{p-value} &
  \multicolumn{2}{c}{change} &
  \multicolumn{1}{c|}{p-value} &
  \multicolumn{2}{c}{change} &
  \multicolumn{1}{c|}{p-value} &
  \multicolumn{2}{c}{change} &
  \multicolumn{1}{c|}{p-value} &
  \multicolumn{2}{c}{change} &
  \multicolumn{1}{c|}{p-value} \\ \hline
Ant. delt. &
  -0.051 & (-17.6\%) &
  \textbf{$<$ 0.001} &
  -0.079 & (-27.2\%) &
  \textbf{$<$ 0.001} &
  -0.113 & (-38.9\%) &
  \textbf{$<$ 0.001} &
  -0.028 & (-11.6\%) &
  \textbf{0.001} &
  -0.062 & (-25.8\%) &
  \textbf{$<$ 0.001} &
  -0.034 & (-16.0\%) &
  \textbf{0.001} \\
Med. delt. &
  0.029 & (10.9\%) &
  0.079 &
  0.010 & (3.7\%) &
  0.432 &
  -0.029 & (-10.8\%) &
  0.091 &
  -0.019 & (-6.5\%) &
  0.121 &
  -0.058 & (-19.6\%) &
  \textbf{$<$ 0.001} &
  -0.039 & (-14.0\%) &
  \textbf{0.005} \\
Post. delt. &
  0.032 & (9.8\%) &
  \textbf{0.027} &
  0.038 & (11.4\%) &
  \textbf{0.024} &
  0.105 & (31.7\%) &
  \textbf{$<$ 0.001} &
  0.005 & (1.4\%) &
  0.667 &
  0.072 & (19.9\%) &
  \textbf{0.004} &
  0.067 & (18.3\%) &
  \textbf{0.009} \\
Biceps &
  0.057 & (16.3\%) &
  \textbf{0.004} &
  0.049 & (14.0\%) &
  \textbf{0.013} &
  0.012 & (3.5\%) &
  0.525 &
  -0.008 & (-1.9\%) &
  0.667 &
  -0.044 & (-10.9\%) &
  \textbf{0.020} &
  -0.037 & (-9.2\%) &
  0.061 \\
Triceps &
  -0.016 & (-4.4\%) &
  0.974 &
  -0.029 & (-7.9\%) &
  0.974 &
  -0.033 & (-8.8\%) &
  0.974 &
  -0.013 & (-3.7\%) &
  0.974 &
  -0.016 & (-4.6\%) &
  0.974 &
  -0.003 & (-1.0\%) &
  0.974 \\
Lat. dorsi &
  0.018 & (5.4\%) &
  0.667 &
  0.021 & (6.5\%) &
  0.643 &
  0.048 & (14.8\%) &
  0.137 &
  0.003 & (1.0\%) &
  0.828 &
  0.030 & (8.9\%) &
  0.667 &
  0.027 & (7.8\%) &
  0.667 \\
Pectoralis &
  -0.098 & (-23.0\%) &
  \textbf{0.008} &
  -0.120 & (-28.3\%) &
  \textbf{0.002} &
  -0.148 & (-34.8\%) &
  \textbf{$<$ 0.001} &
  -0.023 & (-6.9\%) &
  0.179 &
  -0.051 & (-15.4\%) &
  \textbf{0.010} &
  -0.028 & (-9.1\%) &
  0.179 \\
Trapezius &
  -0.099 & (-30.0\%) &
  \textbf{$<$ 0.001} &
  -0.117 & (-35.5\%) &
  \textbf{$<$ 0.001} &
  -0.138 & (-42.0\%) &
  \textbf{$<$ 0.001} &
  -0.018 & (-7.9\%) &
  \textbf{0.140} &
  -0.040 & (-17.1\%) &
  \textbf{0.011} &
  -0.021 & (-10.0\%) &
  0.140 \\ \bottomrule
\end{tabular}%
}
\caption{Pairwise comparisons across conditions for changes in iEMG for the lowering phase. The p-values for significant comparisons ($p<0.05$) are shown in bold. A negative change in iEMG means that the muscle activation was higher in the second condition.}
\label{tab:pairwise_lowering}
\end{table}
\end{landscape}

\newcommand\Tstrut{\rule{0pt}{2.6ex}}         % = `top' strut
\newcommand\Bstrut{\rule[-0.9ex]{0pt}{0pt}}   % = `bottom' strut

\begin{table}[h]
\begin{minipage}[]{0.80\linewidth}
\begin{tabular}{|c|cccccc|}
\hline
Muscle group & pre-no & 25\%-no & 50\%-no & 25\%-pre & 50\%-pre & 50\%-25\%\Tstrut\Bstrut \\
\hline
Ant. delt.  
& \cellcolor[RGB]{157,255,157}\textbf{**} 
& \cellcolor[RGB]{110,255,110}\textbf{**} 
& \cellcolor[RGB]{56,255,56}\textbf{**} 
& \cellcolor[RGB]{201,255,201}\textbf{*} 
& \cellcolor[RGB]{137,255,137}\textbf{**} 
& \cellcolor[RGB]{185,255,185}\textbf{*} \\

Med. delt.  
& \cellcolor[RGB]{255,211,211} 
& \cellcolor[RGB]{255,239,239} 
& \cellcolor[RGB]{195,255,195} 
& \cellcolor[RGB]{231,255,231} 
& \cellcolor[RGB]{157,255,157}\textbf{**} 
& \cellcolor[RGB]{179,255,179}\textbf{*} \\

Post. delt. 
& \cellcolor[RGB]{255,211,211} 
& \cellcolor[RGB]{255,211,211} 
& \cellcolor[RGB]{255,151,151}\textbf{*} 
& \cellcolor[RGB]{255,255,255} 
& \cellcolor[RGB]{255,201,201} 
& \cellcolor[RGB]{255,201,201} \\

Biceps      
& \cellcolor[RGB]{255,177,177}\textbf{*} 
& \cellcolor[RGB]{255,173,173}\textbf{*} 
& \cellcolor[RGB]{255,235,235} 
& \cellcolor[RGB]{255,253,253} 
& \cellcolor[RGB]{203,255,203} 
& \cellcolor[RGB]{201,255,201} \\

Triceps     
& \cellcolor[RGB]{239,255,239} 
& \cellcolor[RGB]{217,255,217} 
& \cellcolor[RGB]{189,255,189} 
& \cellcolor[RGB]{233,255,233} 
& \cellcolor[RGB]{205,255,205} 
& \cellcolor[RGB]{225,255,225} \\

Trapezius   
& \cellcolor[RGB]{137,255,137}\textbf{**} 
& \cellcolor[RGB]{102,255,102}\textbf{**} 
& \cellcolor[RGB]{22,255,22}\textbf{**} 
& \cellcolor[RGB]{215,255,215}\textbf{*}
& \cellcolor[RGB]{112,255,112}\textbf{**} 
& \cellcolor[RGB]{147,255,147}\textbf{**} \\

Lat. dorsi  
& \cellcolor[RGB]{255,253,253} 
& \cellcolor[RGB]{255,243,243} 
& \cellcolor[RGB]{241,255,241} 
& \cellcolor[RGB]{255,245,245} 
& \cellcolor[RGB]{239,255,239} 
& \cellcolor[RGB]{229,255,229} \\

Pectoralis major  
& \cellcolor[RGB]{131,255,131}\textbf{**} 
& \cellcolor[RGB]{98,255,98}\textbf{**} 
& \cellcolor[RGB]{42,255,42}\textbf{**} 
& \cellcolor[RGB]{217,255,217} 
& \cellcolor[RGB]{145,255,145}\textbf{**} 
& \cellcolor[RGB]{179,255,179}\textbf{*} \\

\hline
\end{tabular}
\end{minipage}
\begin{minipage}[]{0.1\linewidth}
\includegraphics[height=4cm]{BardiTableColorbar.pdf}
\end{minipage}
\caption{\label{tab:iEMG_deltas_visual}EMG statistical results for the pairwise post-hoc tests comparing changes in muscular activity across all muscle groups for the entire arm cycle. The following significance codes are used for the p-values: \textbf{**} = [0, 0.001), \textbf{*} = [0.001, 0.05), and a blank space represents [0.05, 1]. The percentage changes in normalized iEMG, relative to the first condition being compared, is coded in color, with green representing a decrease in iEMG from the second to the first condition, and red representing an increase.}
\end{table}